\documentclass[sigconf]{acmart}
\AtBeginDocument{%
  \providecommand\BibTeX{{%
    \normalfont B\kern-0.5em{\scshape i\kern-0.25em b}\kern-0.8em\TeX}}}

\copyrightyear{2022}
\acmYear{2022}
\setcopyright{acmcopyright}\acmConference[WWW '22]{Proceedings of the ACM Web Conference 2022}{April 25--29, 2022}{Virtual Event, Lyon, France}
\acmBooktitle{Proceedings of the ACM Web Conference 2022 (WWW '22), April 25--29, 2022, Virtual Event, Lyon, France}
\acmPrice{15.00}
\acmDOI{10.1145/3485447.3511986}
\acmISBN{978-1-4503-9096-5/22/04}

\usepackage{bm}
\usepackage{color, soul}
\usepackage{subfigure}
\usepackage{algorithm}
\usepackage{algorithmic}

\usepackage{listings}

\usepackage{xspace}
\newcommand{\sys}{\textsc{PaSca}\xspace}

\newcommand{\para}[1]{{\vspace{2pt} \bf \noindent #1 \hspace{1pt}}}
\usepackage{multirow}
\usepackage{makecell}

\begin{document}

\title{\sys: a Graph Neural Architecture Search System under the Scalable Paradigm}

\author{Wentao Zhang$^{1,2}$, Yu Shen$^1$, Zheyu Lin$^1$, Yang Li$^1$, Xiaosen Li$^2$, Wen Ouyang$^2$,\\Yangyu Tao$^2$, Zhi Yang$^{1}$, Bin Cui$^{1,3}$}
\affiliation{
$^1$School of CS \& Key Lab of High Confidence Software Technologies, Peking University\\$^2$Tencent Inc.~~~~~$^3$Institute of Computational Social Science, Peking University (Qingdao), China\country{}}
\affiliation{
$^1$\{wentao.zhang, shenyu, linzheyu, liyang.cs, yangzhi, bin.cui\}@pku.edu.cn \\$^2$\{wtaozhang, hansenli, gdpouyang, brucetao\}@tencent.com\country{}
}

\renewcommand{\shortauthors}{WT Zhang, et al.}
\renewcommand{\authors}{WT Zhang, Y Shen, ZY Lin, Y Li, XS Li, W Ouyang, YY Tao, Z Yang, and B Cui}
\renewcommand{\shortauthors}{Wentao Zhang, et al.}

\begin{abstract}
Graph neural networks (GNNs) have achieved state-of-the-art performance in various graph-based tasks.
However, as mainstream GNNs are designed based on the neural message passing mechanism, they do not scale well to data size and message passing steps.
Although there has been an emerging interest in the design of scalable GNNs, current researches focus on specific GNN design, rather than the general \emph{design space}, limiting the discovery of potential scalable GNN models.
This paper proposes \sys, a new paradigm and system that offers a principled approach to systemically construct and explore the design space for scalable GNNs, rather than studying individual designs.
Through deconstructing the message passing mechanism, \sys presents a novel Scalable Graph Neural Architecture Paradigm (SGAP), together with a general architecture design space consisting of 150k different designs. 
Following the paradigm, we implement an auto-search engine that can automatically search well-performing and scalable GNN architectures to balance the trade-off between multiple criteria (e.g., accuracy and efficiency) via multi-objective optimization. 
Empirical studies on ten benchmark datasets demonstrate that the representative instances (i.e., \sys-V1, V2, and V3) discovered by our system achieve consistent performance among competitive baselines.
Concretely, \sys-V3 outperforms the state-of-the-art GNN method JK-Net by 0.4\% in terms of predictive accuracy on our large industry dataset while achieving up to $28.3\times$ training speedups.
\end{abstract}



\maketitle

\section{Introduction}
\label{intro}
Graph neural networks (GNNs)~\cite{wu2020comprehensive} have become the state-of-the-art methods in many graph representation learning scenarios such as node classification~\cite{miao2021degnn, zhang2021rod, ma2021improving, dong2021equivalence}, link prediction~\cite{DBLP:journals/chinaf/GuoQX021, wu2021hashing, wang2021graph, cai2020multi}, recommendation~\cite{ying2018graph, monti2017geometric, wu2020graph, hsu2021retagnn}, and knowledge graphs~\cite{wang2018deep, DBLP:conf/www/WangWSWNAXYC21, wang2018zero, bastos2021recon}. 
Most GNN pipelines can be described in terms of the neural message passing (NMP) framework~\cite{hamilton2017inductive}, which is based on the core idea of recursive neighborhood aggregation and transformation.
Specifically, during each iteration, the representation of each node is updated (with neural networks) based on messages received from its neighbors. 
Since they typically need to perform a recursive neighborhood expansion to gather neural messages repeatedly, this process leads to an expensive neighborhood expansion, which grows exponentially with layers. The exponential growth of
neighborhood size corresponds to an exponential IO overhead, which is the major challenge of large-scale GNN computation.

\begin{figure}
\vspace{-4mm}
 \centering
 \subfigure[Speedup]{
  \scalebox{0.47}[0.47]{
   \includegraphics[width=1\linewidth]{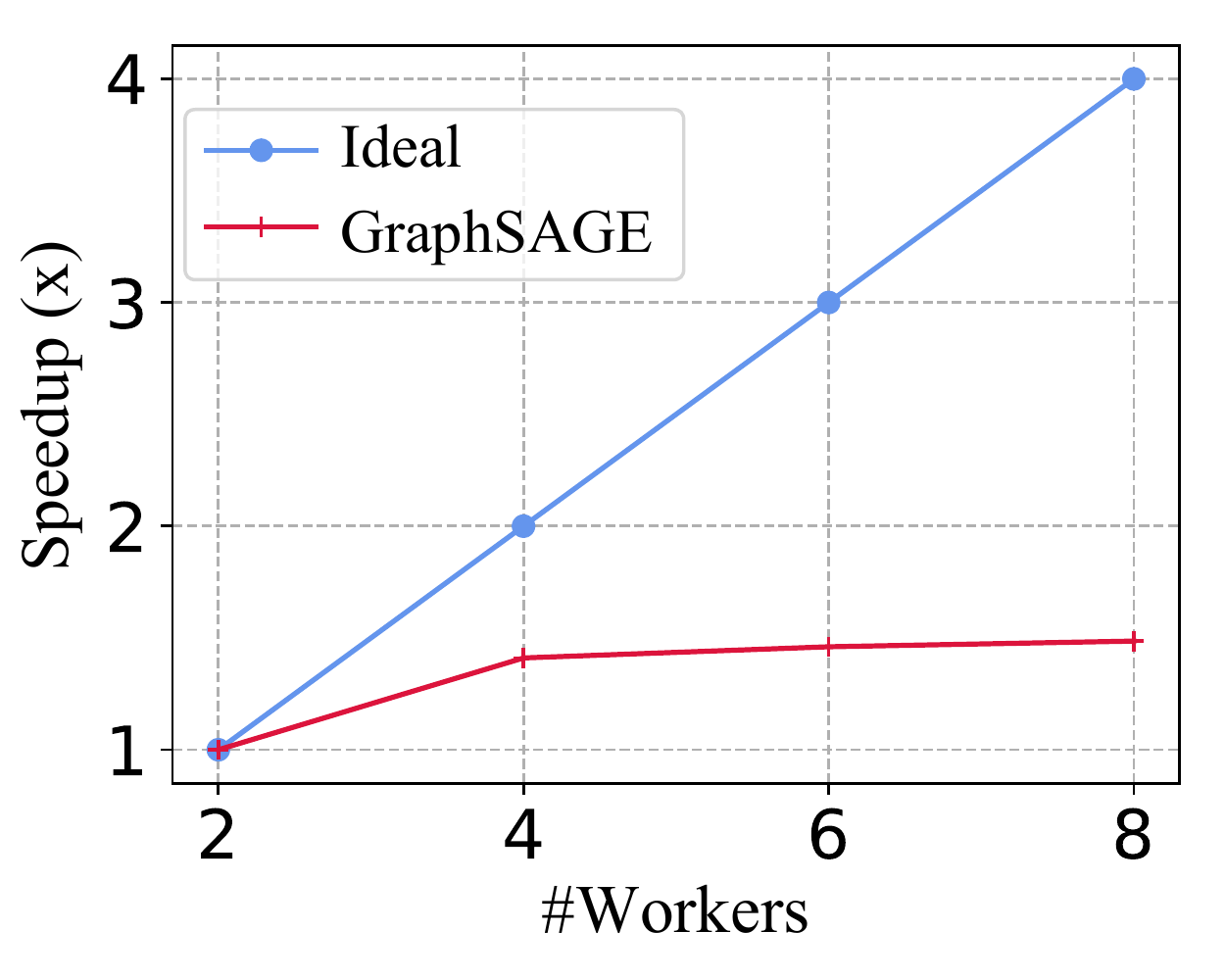}
 }}
 \subfigure[Bottleneck]{
  \scalebox{0.45}[0.45]{
   \includegraphics[width=1\linewidth]{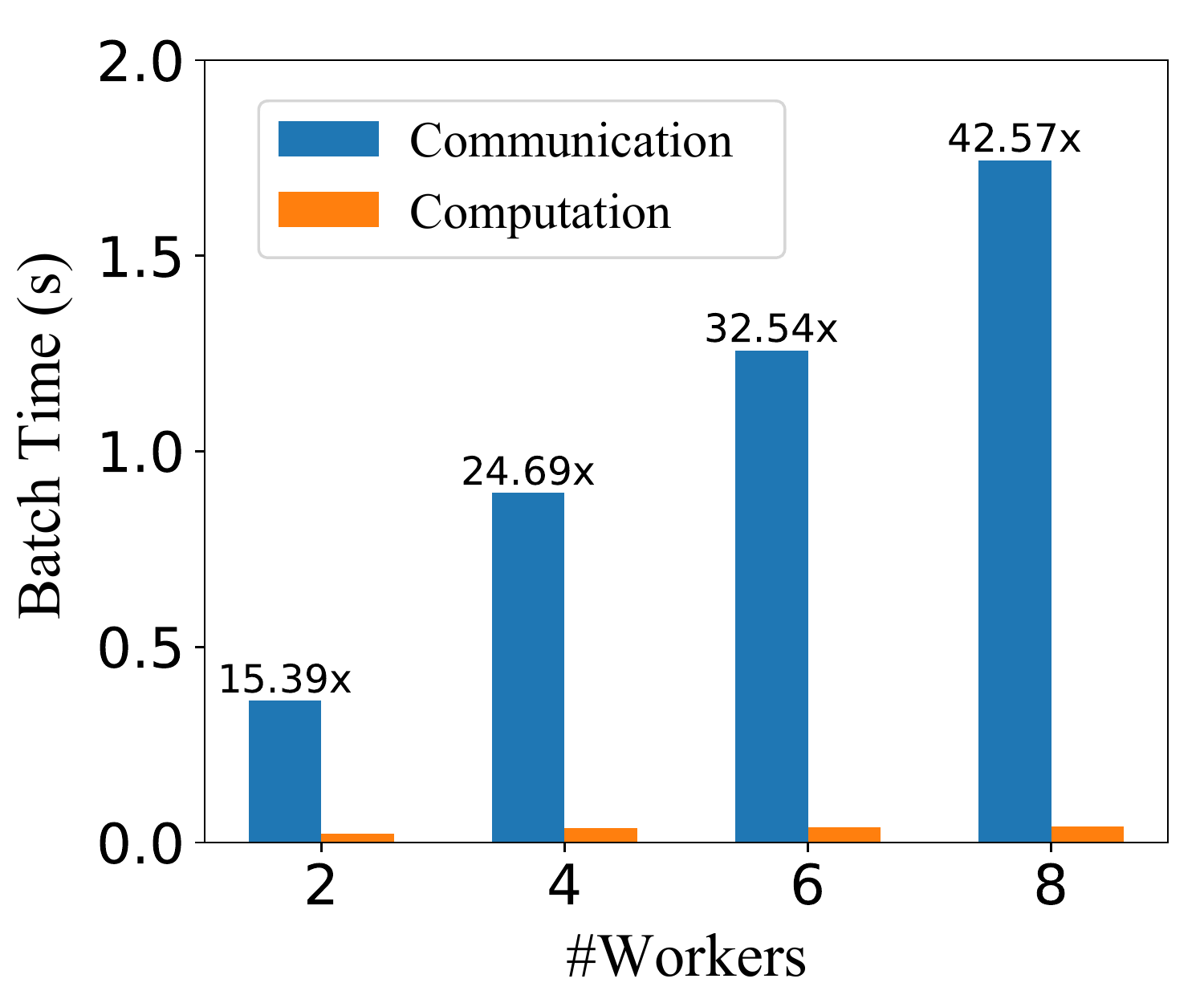}
 }}
\vspace{-4mm}
\caption{The speedup and bottleneck of a two-layer GraphSAGE along with the increased workers on Reddit dataset. }
\vspace{-7mm}
\label{fig:dglunscale}
\end{figure}

To scale up GNNs to web-scale graphs, recent work focuses on designing training frameworks with sampling approaches (e.g., DistDGL~\cite{zheng2020distdgl}, NextDoor~\cite{jangda2021accelerating}, SeaStar~\cite{wu2021seastar}, FlexGraph~\cite{wang2021flexgraph}, Dorylus~\cite{thorpe2021dorylus},
GNNAdvisor~\cite{wang2021gnnadvisor}, etc.). 
Although distributed training is applied in these frameworks, they still suffer from high communication costs due to the recursive neighborhood aggregation during the training process.
To demonstrate this issue, we utilize distributed training functions provided by DGL~\cite{bronstein2017geometric} to execute the train pipeline of GraphSAGE~\cite{hamilton2017inductive}. 
We partition the Reddit dataset across multiple machines and treat each GPU as a worker, and then calculate the speedup relative to the runtime of two workers.
Figure~\ref{fig:dglunscale} illustrates the training speedup along with the number of workers and the bottleneck in distributed settings.
In particular, Figure~\ref{fig:dglunscale}(a) shows that the scalability of GraphSAGE is limited even when the mini-batch training and graph sampling method are adopted. 
Figure~\ref{fig:dglunscale}(b) further shows that the scalability is mainly bottlenecked by the aggregation procedure in which high data loading cost is incorporated to gather neighborhood information. 

Different from the recently developed GNN systems~\cite{thorpe2021dorylus,wang2021gnnadvisor}, we address the scalability challenges from an orthogonal perspective: re-designing the GNN pipeline to make the computing naturally scalable. 
To ensure scalability, we consider a different GNN training pipeline from most existing work: treating data aggregation over the graph as pre/post-processing stages that are separate from training. 
While there has been an emerging interest in specific architectural designs with decoupled pipeline~\cite{wu2019simplifying,sign_icml_grl2020,zhu2021simple, zhang2021node}, current researches only focus on \emph{specific GNN instances}, rather than the \emph{general design space}, which limits the discovery of potential scalable GNN variants.
In addition, new architecture search systems are required to perform extensive exploration over the design space for scalable GNNs, which is also a major motivation to our work.

To the best of our knowledge, we propose the first paradigm and system -- \sys to explore the designs of scalable GNN, which makes the following contributions:

\para{Scalable Paradigm.} We introduce the Scalable Graph Neural Architecture Paradigm (SGAP) with three operation abstractions: (1) \texttt{graph\_aggregator} captures the structural
information via graph aggregation operations, (2) \texttt{message\_aggregator} combines different levels of structural information, and (3) \texttt{message\_updater} generates the prediction based on the multi-scale features. 
Compared with the recently published scalable GNN systems, the SGAP interfaces in \sys are motivated and implemented differently: (1) The APIs of GNN systems are used to express existing GNNs, whereas we propose a novel GNN pipeline abstraction to define the general design space for scalable GNN architectures. (2) The existing system contains two stages --- sampling and training, where sampling is not a decoupled pre-processing stage and needs to be performed for each training iteration. 
By contrast, the SGAP paradigm considers propagation as pre/post-processing and does not require the expensive neighborhood expansion during training.

\begin{figure}[tpb]
\vspace{-4mm}
    \centering
    \includegraphics[width=0.85\linewidth]{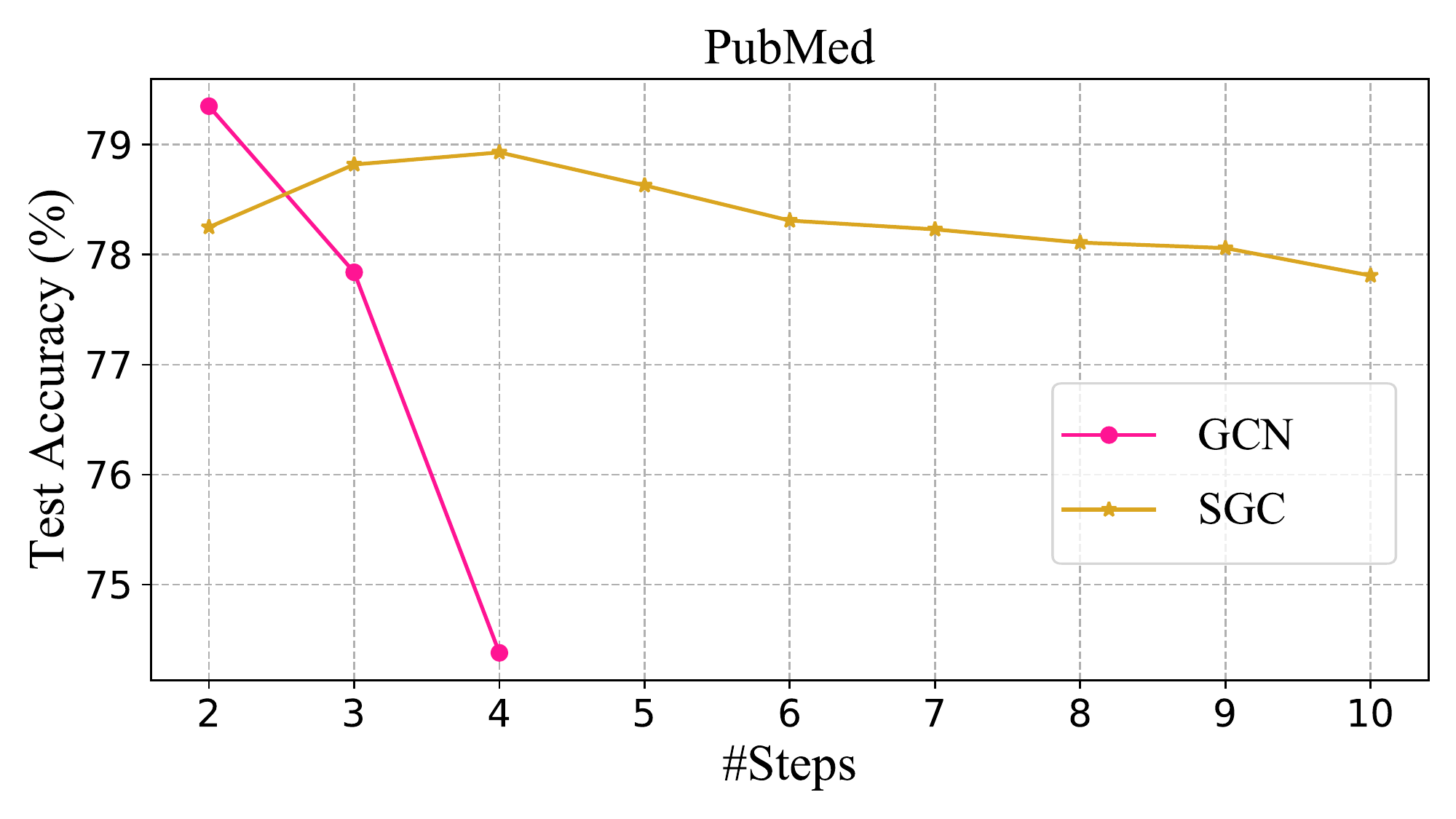}
    \vspace{-4mm}
    \caption{Test accuracy of different models along with the increased aggregation steps on PubMed dataset.}
    \label{fig:depth}
    \vspace{-7mm}
\end{figure}

\para{Design Space.}Based on the proposed SGAP paradigm, we further propose a general design space consisting of 6 design dimensions, resulting in 150k possible designs of scalable GNN. 
We find that recently emerging scalable GNN models, such as SGC~\cite{wu2019simplifying}, SIGN~\cite{sign_icml_grl2020}, S$^2$GC~\cite{zhu2021simple} and GBP~\cite{chen2020scalable} are special instances in our design space. 
Instead of simply generalizing existing specific GNN designs, we propose a design space with adaptive aggregation and a complementary post-processing stage beyond what is typically considered in the literature.
The extension is motivated by the observation that previous GNNs (e.g., GCN~\cite{kipf2017semi} and SGC) suffer from model
scalability issue, as shown in Figure~\ref{fig:depth}. 
Here we use \emph{model scalability} to describe its capability to cope with the large-scale neighborhood with increased aggregation step $k$.
We find the underlying reason is that their aggregation processes are restricted to a fixed-hop neighborhood and are insensitive to the actual demands of different nodes, which may lead to two limitations preventing them from unleashing their full potential: 
(1) long-range dependencies cannot be fully leveraged due to limited hops/layers, and (2) local information are lost due to the introduction of irrelevant nodes and unnecessary messages when the number of hops increases (i.e., over-smoothing issue ~\cite{li2018deeper,zhang2021evaluating, miao2021lasagne}). 
Through extending design space with adaptive aggregators, we could discover new GNNs to achieve both model scalability and training scalability.

\para{Auto-search Engines.} We design and implement a search system that automates the search procedure for well-performing scalable GNN architectures to explore the proposed design space instead of the manual design. 
Our search system contains the following two engines: 
(1) \emph{Suggestion engine} that implements a multi-objective search algorithm, which aims to find Pareto-optimal GNN instances given multiple criteria (e.g., predictive performance, inference time, resource consumption), allowing for a designer to select the best Pareto-optimal solution based on specific requirements; 
(2) \emph{Evaluation engine} that evaluates the GNN instances from the search engine in a distributed manner. 
Due to the repetitive expansion in the training stage of GNNs, it is hard for existing GNN systems to scale to increasing workers. 
Based on the SGAP paradigm, the evaluation engine in \sys involves the expensive neighborhood expansion only once in the pre/post-processing stages, and thus ensuring the scalability upon the number of training workers.
To support the new pipeline, we implement two components: (1) the distributed graph data aggregator to pre/post-process data over graph structure, and (2) the distributed trainer where workers only need to exchange neural parameters.

Based on our auto-search system \sys, we discover new scalable GNN instances from the proposed design space for different accuracy-efficiency requirements.
Extensive experiments on ten graph datasets demonstrate the superior training scalability/efficiency and performance of searched representatives given by \sys among competitive baselines. 
Concretely, the representatives (i.e., \sys-V2 and \sys-V3) outperform the state-of-the-art JK-Net by 0.2\% and 0.4\% in predictive accuracy on our industry dataset, while achieving up to $56.6\times$ and $28.3\times$ training speedups, respectively.

\para{Relevance to Web.}GNNs have recently been applied to a broad spectrum of web
research such as social influence prediction~\cite{qiu2018deepinf, DBLP:conf/www/PiaoZXCL21}, network
role discovery~\cite{centrality2018, rozemberczki2021pathfinder}, recommendation system~\cite{ying2018graph, hsu2021retagnn, wu2021hashing}, and fraud/spam detection~\cite{spam2019,liu2021pick}. However, scalability is a major challenge that precludes GNN-based methods in practical web-scale graphs.
Moreover, manually designing the well-behaved GNNs for web tasks requires immense human expertise. 
To bridge this gap, we highlight the relevance of the proposed \sys platform to GNN-based web research.
First, \sys provides easier support for experts in solving their web problems via scalable GNN paradigm. Domain experts only need to provide properly formatted datasets, and \sys can automatically 
search suitable and scalable GNN designs to the web-scale graphs. 
So \sys permits a transition from particular GNN instances to GNN design space, which offers exciting opportunities for scalable GNN architecture innovation.
Second, \sys provides the dis-aggregated execution pipeline for efficiently training and evaluating searched GNN models, without resorting to any approximation technique (e.g., graph sampling). 
Given these advancements in scalability and automaticity, our \sys system enables practical and scalable GNN-based implementation for web-scale tasks, and thus significantly reducing the barrier
when applying GNN models in web research. 

\section{Preliminary}
\para{GNN Pipelines.}
\label{mp}
Considering a graph $\mathcal{G=(V, E)}$ with nodes $\mathcal{V}$, edges $\mathcal{E}$ and features for all nodes $\mathbf{x}_v \in \mathbb{R}^d, \forall v \in \mathcal{V}$, Most GNNs can be formulated using the neural message passing framework, like GCN~\cite{kipf2017semi}, GraphSAGE~\cite{hamilton2017inductive}, GAT~\cite{velivckovic2018graph}, and GraphSAINT~\cite{zeng2020graphsaint}, where each layer adopts one aggregation and an update function.
At time step $t$, a message vector $\mathbf{m}^{t}_v$ is computed with the representations of its neighbors $\mathcal{N}_v$ using an aggregate function, and $\mathbf{m}^{t}_v$ is then updated by a neural-network based update function:
\begin{equation}
\begin{aligned}
& \mathbf{m}^t_v \gets \texttt{aggregate}\left(\left\{\mathbf{h}^{t-1}_u|{u \in \mathcal{N}_v}\right\}\right),\ \mathbf{h}^t_v \gets \texttt{update}(\mathbf{m}^t_v).
\end{aligned}
\label{eq:mp}
\end{equation}
In this way, messages are passed for $K$ time steps in a $K$-layer GNN so that the steps of message passing correspond to the GNN depth. Taking the vanilla GCN~\cite{kipf2017semi} as an example, we have:
\begin{equation}
\begin{aligned}
 &\texttt{GCN-aggregate}\left(\left\{\mathbf{h}^{t-1}_u|{u \in \mathcal{N}_v}\right\}\right)=\sum_{u \in \mathcal{N}_v}\mathbf{h}_u^{t-1}/\sqrt{\tilde{d}_v\tilde{d}_u},\\
  &\texttt{GCN-update}(\mathbf{m}^t_v)=\sigma(W\mathbf{m}^t_v),\nonumber
 \end{aligned}
\end{equation}
where $\tilde{d}_v$ is the degree of node $v$ obtained from the adjacency matrix with self-connections $\tilde{A}=I+A$. Recently, some GNN variants adopt the 
decoupled neural message passing (DNMP) for better graph learning. More details can be found in Appendix~\ref{DNMP}.

\para{Scalable GNN Instances.}
Following SGC~\cite{wu2019simplifying}, a recent direction for scalable GNN is to remove the non-linearity between each layer in the forward aggregation, and models in this direction have achieved state-of-the-art performance in leaderboards of Open Graph Benchmark~\cite{hu2020open}. 
Concretely, SIGN~\cite{sign_icml_grl2020} proposes to concatenate different iterations of aggregated feature messages, while S$^2$GC~\cite{zhu2021simple} proposes a simple spectral graph convolution to average them.
In addition, GBP~\cite{chen2020scalable} applies constants to weight aggregated feature messages of different layers.
As current researches focus on studying specific architectural designs, we systematically study the architectural design space for scalable GNNs.

\para{Graph Neural Architecture Search.}
As a popular direction of AutoML~\cite{li2021volcanoml,li2021mfes,he2021automl}, neural architecture search~\cite{fang2021eat, DBLP:conf/ijcai/ZhangW021,shen2021proxybo} has been proposed to solve the labor-intensive problem of neural architecture design.
Auto-GNN~\cite{zhou2019auto} and GraphNAS~\cite{gao2019graphnas} are early approaches that apply reinforcement learning with RNN controllers on a fixed search space. 
You et al. ~\cite{you2020design} define a similarity metric and search for the best transferable architectures across tasks via random search. 
Based on DARTS~\cite{liu2019darts}, GNAS~\cite{cai2021rethinking} proposes the differentiable searching strategy to search for GNNs with optimal message-passing step. 
DSS~\cite{li2021one} also adopts the differentiable strategy but optimizes over a dynamically updated search space.
\sys differs from these works in two aspects: 
(1) To pursue efficiency and scalability on large graphs, \sys searches for scalable architectures under the novel SGAP paradigm instead of classic architectures under the message passing framework; 
(2) Rather than optimizing the predictive performance alone, \sys tackles the accuracy-efficiency trade-off through multi-objective optimization, and provides architectures to meet different needs of performance and inference time.

\section{\sys Abstraction}
To address data and model scalability issues mentioned in Section~\ref{intro}, we propose a novel abstraction under which more scalable GNNs can be derived. 
Then we define the general design space for scalable GNNs based on the proposed abstraction.

\subsection{SGAP Paradigm}
\label{paradigm}
Our \sys system introduces a Scalable Graph Neural Architecture
Paradigm (SGAP) for designing scalable GNNs. As shown in Algorithm~\ref{alg:Pasca}, it consists of the following three decoupled stages:  

\para{Pre-processing.} For each node $v$, we range the step $t$ from $1$ to $K_{pre}$, where $K_{pre}$ is the maximum feature aggregation step. 
At each step $t$, we use an operator, namely \texttt{graph\_aggregator}, to aggregate the message vector collected from the neighbors $\mathcal{N}_v$:
\begin{equation}
\mathbf{m}^t_v \gets \texttt{graph\_aggregator}\left(\left\{\mathbf{m}^{t-1}_u|{u \in \mathcal{N}_v}\right\}\right),
\end{equation}
where $\mathbf{m}^0_v=\mathbf{x}_v$. 
The messages are passed for $K_{pre}$ steps in total during pre-processing, and $\mathbf{m}^t_v$ at step $t$ can gather the neighborhood information from nodes that are $t$-hop away (lines 4-9). 

\begin{algorithm}[t]
  \caption{Scalable graph neural architecture paradigm. 
  }
  \label{alg:Pasca}
\begin{algorithmic}[1]
  \REQUIRE Graph $\mathcal{G=(V, E)}$, maximum aggregation steps for pre-processing and post-processing $K_{pre}, K_{post}$, feature $\mathbf{x}_v$.
  \ENSURE Prediction message $\mathbf{m}^{K_{post}}_v$, $\forall v \in \mathcal{V}$.
   \STATE Initialize message set $\mathcal{M}_v=\{\mathbf{x}_v\}$, $ \forall v \in \mathcal{V}$;\\
   \STATE // \textbf{Stage 1: Pre-processing}\\
   \STATE Initialize feature message $\mathbf{m}_v^0=\mathbf{x}_v$, $\forall v \in \mathcal{V}$;\\
  \FOR{$1\leq t\leq K_{pre}$}
    \FOR{$v\in \mathcal{V}$}
        \STATE $\mathbf{m}^t_v \gets$ \texttt{graph\_aggregator}$(\mathbf{m}^{t-1}_{\mathcal{N}_v})$;\\
        \STATE $\mathcal{M}_v = \mathcal{M}_v \cup \{\mathbf{m}^t_v\}$;\\
    \ENDFOR
   \ENDFOR
    \STATE // \textbf{Stage 2: Model-training}
    \FOR{$v\in \mathcal{V}$}
    \STATE $\mathbf{c}_v \gets $\texttt{message\_aggregator}$(\mathcal{M}_v)$;\\
    \STATE $\mathbf{h}_v \gets $\texttt{message\_updater}$(\mathbf{c}_v)$;\\ 
    \ENDFOR
  \STATE // \textbf{Stage 3: Post-processing}\\
   \STATE Initialize feature message $\mathbf{m}_v^0=\mathbf{h}_v$, $\forall v \in \mathcal{V}$;\\
  \FOR{$1\leq t\leq K_{post}$}
  
    \FOR{$v\in \mathcal{V}$}
        \STATE $\mathbf{m}^t_v \gets$ \texttt{graph\_aggregator}$(\mathbf{m}^{t-1}_{\mathcal{N}_v})$;\\
    \ENDFOR
  \ENDFOR
  \RETURN $\mathbf{m}^{K_{post}}_v$, $\forall v \in \mathcal{V}$;
\end{algorithmic}
\end{algorithm}

\para{Model-training.}
The multi-hop messages $\mathcal{M}_v=\{\mathbf{m}^{t}_v\ |\ 0\leq t \leq K_{pre}\}$ are then aggregated into a single combined message vector $\mathbf{c}_v$ for each node $v$ (line 12) as:
\begin{equation}
\mathbf{c}_v \gets \texttt{message\_aggregator}(\mathcal{M}_v).
\end{equation}

Note that, if the \texttt{message\_aggregator} is not applied, the combined message vector $\mathbf{c}_v$ is set to the message of the last step $\mathbf{m}^{K_{pre}}_v$. 

We then use the $\texttt{message\_updater}$ to learn the class distribution of all nodes, i.e., the soft predictions (softmax outputs) predicted by the updater (line 13). Specifically, \sys applies the Multi-layer Perceptron (MLP) as the updater, and we denote the depth of MLP as the transformation step $K_{trans}$. It learns node embedding $\mathbf{h}_v$ from the combined message vector $\mathbf{c}_v$:
\begin{equation}
\mathbf{h}_v \gets \texttt{message\_updater}(\mathbf{c}_v).
\end{equation}

\para{Post-processing.}Motivated by Label Propagation~\cite{wang2007label} which aggregates the node labels, we regard the soft predictions as new features (line 16). Then, we use the $\texttt{graph\_aggregator}$ again at each step to aggregate the adjacent node predictions and make the final prediction (lines 17-21) as:
\begin{equation}
\mathbf{m}^t_v \gets \texttt{graph\_aggregator}\left(\left\{\mathbf{m}^{t-1}_u|{u \in \mathcal{N}_v}\right\}\right),
\end{equation}
where $\mathbf{m}^0_v=\mathbf{h}_v$ is the original node prediction.

We introduce SGAP to address both training and model scalability challenges. Specifically, it differs from the previous NMP and DNMP framework in terms of message type, message scale, and pipeline: (1)
To perform the aggregate function for the next step, existing GNNs in NMP and DNMP update the hidden state $\mathbf{h}^{t}_v$ by applying the message vector $\mathbf{m}^{t}_v$ with neural networks.
By contrast, SGAP allows passing node feature messages without applying \texttt{graph\_aggregator} on the hidden states. 
As a result, this message passing procedure is independent of learnable model parameters and can be easily pre-computed, thus leading to high scalability and speedups. 
(2) Most GNNs in NMP and DNMP only utilizes the last message vector $\mathbf{m}^{K_{pre}}_v$ to compute the final hidden state $\mathbf{h}^{K_{pre}}_v$. 
SGAP assumes that the optimal neighborhood expansion size should be different for each node $v$, and thus we retain all the messages $\{\mathbf{m}^t_v|t\in[1,\ K_{pre}]\}$ that a node $v$ receives over different steps (i.e., localities). 
The multi-scale messages are then aggregated per node into a single vector via \texttt{message\_aggregator}, such that we could balance the preservation of information from both local and extended (multi-hop) neighborhoods for each node. 
(3) Besides feature aggregation, we propose a complementary
post-processing stage to aggregate predictions (soft labels), which is not typically considered in the existing literature.

\begin{table*}[h]
\vspace{-3mm}
\caption{ The search space for scalable GNNs in our \sys system.}
\vspace{-3mm}
\centering
{
\noindent
\renewcommand{\multirowsetup}{\centering}
\resizebox{0.85\linewidth}{!}{
\begin{tabular}{ccccc}
\toprule
\textbf{Stages}& \textbf{Name}&\textbf{Range/Choices}&\textbf{Type}\\
\midrule
\multirowcell{2}{Pre-processing} & Aggregation steps ($K_{pre}$) & [0, 10] & Integer\\
&Graph aggregators ($GA_{pre}$) & \{Aug.NA, PPR($\alpha$ = 0.1), PPR($\alpha$ = 0.2), PPR($\alpha$ = 0.3), Triangle. IA\}& Categorical\\
\midrule
\multirowcell{2}{Model training}&Message aggregators ($MA$)& \{None, Mean, Max, Concatenate, Weighted, Adaptive\}& Categorical\\
&Transformation steps ($K_{trans}$) & [1, 10] & Integer\\
\midrule
\multirowcell{2}{Post-processing}& Aggregation steps ($K_{post}$) & [0, 10] & Integer\\
&Graph aggregators ($GA_{post}$) & \{Aug.NA, PPR($\alpha$ = 0.1), PPR($\alpha$ = 0.2), PPR($\alpha$ = 0.3), Triangle. IA\}& Categorical\\
\bottomrule
\end{tabular}}}
\vspace{-2mm}
\label{tab:search_space}
\end{table*}

\subsection{Design Space under SGAP}
Following the SGAP paradigm, we propose a general design space for scalable GNNs, as shown in Table~\ref{tab:search_space}.
The design space contains three integer and three categorical parameters, which are responsible for the choice of aggregators and the steps of aggregation and transformation. 
Each configuration sampled from the search space represents a unique scalable architecture, resulting in 150k possible designs in total. 
One can also include more aggregators in the current space with future state-of-the-arts.
In the following, we first introduce the aggregators used in our design space, and then explore interesting GNN instances in our defined space.

\subsubsection{Graph Aggregators}\label{sec:ga}
To capture the information of nodes that are several hops away, \sys adopts a \texttt{graph\_aggregator} to combine the nodes with their neighbors during each time step. 
Intuitively, it is unsuitable to use a fixed \texttt{graph\_aggregator} for each task since the choice of graph aggregators depends on the graph structure and features. Thus \sys provides three different graph aggregators to cope with different scenarios, and one could add more aggregators following the semantic of \texttt{graph\_aggregator}.

\para{Augmented normalized adjacency (Aug. NA)~\cite{kipf2017semi}.}It applies the random walk normalization on the augmented adjacency matrix $\tilde{A}=I+A$, which is simple yet effective on a range of GNNs. The normalized \texttt{graph\_aggregator} is: 
\begin{small}
\begin{equation}
\mathbf{m}_v^{t}=\sum_{u \in \mathcal{N}_v}\frac{1}{\tilde{d}_u}\mathbf{m}_u^{t-1}.
\end{equation}
\end{small}

\para{Personalized PageRank (PPR).}  It focuses on its local neighborhood using a restart probability $\alpha \in \left(0,1 \right]$ and performs well on graphs with noisy connectivity. While the calculation of the fully personalized PageRank matrix is computationally expensive, we apply its approximate computation ~\cite{klicpera2019predict}:

\begin{small}
\begin{equation}
    \mathbf{m}_v^{t}=\alpha \mathbf{m}_v^0+(1-\alpha)\sum_{u \in \mathcal{N}_v}\frac{1}{\sqrt{\tilde{d}_v\tilde{d}_u}}\mathbf{m}_u^{t-1},
\end{equation}
\end{small}
where the restart probability $\alpha$ allows to balance preserving locality (i.e., staying close to the root node to avoid over-smoothing) and leveraging the information from a large neighborhood.

\para{Triangle-induced adjacency (Triangle. IA)~\cite{monti2018motifnet}.}It accounts for the higher-order structures and helps distinguish strong and weak ties on complex graphs like social graphs. We assign each edge a weight representing the number of different triangles it belongs to, which forms a weight matrix $A^{tri}$. We denote $d^{tri}_v$ as the degree of node $v$ from the weighted adjacency matrix $A^{tri}$. The aggregator is then calculated by applying a row-wise normalization:
\begin{small}
\begin{equation}
    \mathbf{m}_v^{t}=\sum_{u \in \mathcal{N}_v}\frac{1}{d_v^{tri}}\mathbf{m}_u^{t-1}.
\end{equation}
\end{small}

\subsubsection{Message Aggregators}
Before updating the hidden state of each node, \sys proposes to apply a \texttt{message\_aggregator} to combine messages obtained by \texttt{graph\_aggregator} per node into a single vector, such that the subsequent model learns from the multi-scale neighborhood of a given node. 
We summarize the different message aggregators \sys as follows,

\para{Non-adpative aggregator.}This type of aggregator does not consider the correlation between messages and the center node. The messages are directly concatenated or summed up with weights to obtain the combined message vector as, 
\begin{equation}
\small
    c_{msg} \gets \oplus_{\mathbf{m}_v^i\in M_v}w_if(\mathbf{m}_v^i),
\end{equation}
where $f$ is a function used to reduce the dimension of message vectors, and $\oplus$ can be concatenating or pooling operators including average pooling or max pooling.
Note that, for aggregator type ``Mean'', ``Max'' and ``Concatenate'', each weight $w_i$ is set to 1 for each message. For aggregator type ``Weighted'', we set the weight to constants following GBP~\cite{chen2020scalable}.
Compared with pooling operators, though the concatenating operator keeps all the input message information, the dimension of its outputs increases as $K_{pre}$ grows, leading to additional computational cost in the downstream updater.

\begin{figure}
    \centering
    \includegraphics[width=\linewidth]{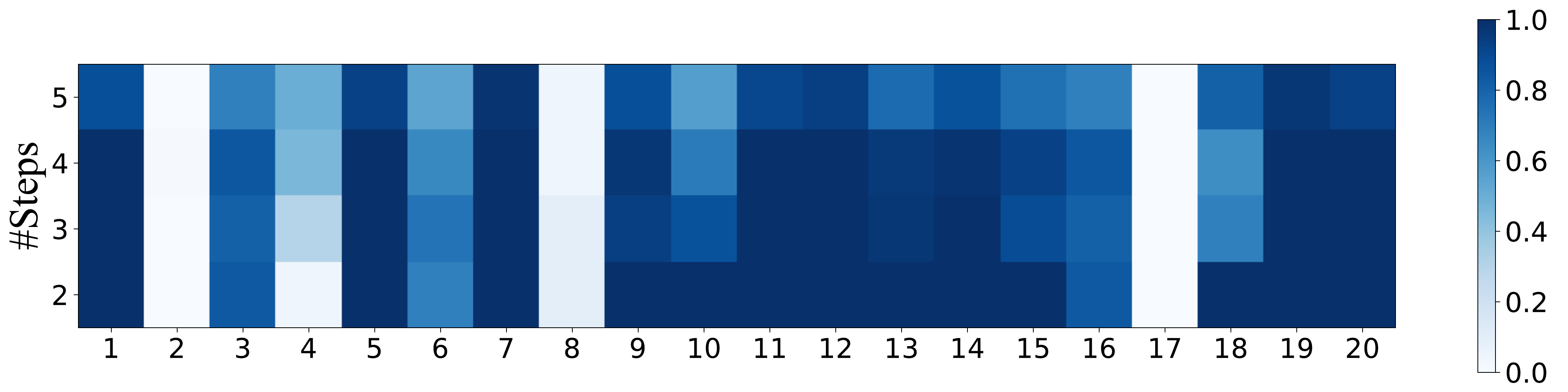}
    \vspace{-6mm}
    \caption{The influence of aggregation steps on 20 randomly sampled nodes on Citeseer dataset.}
    \vspace{-3mm}
    \label{fig:flexibility}
\end{figure}

\para{Adpative aggregators.}The messages of different hops make different contributions to the final performance.
As shown in Figure \ref{fig:flexibility}, we apply GCN with different layers to conduct node classification on Citeseer. Note that the X-axis is the node id and Y-axis is the aggregation steps (number of layers in GCN). The color from white to blue represents the ratio of being predicted correctly in 50 different runs.
We observe that most nodes are well classified with two steps, and as a result, most carefully designed GNN models are set with two layers (i.e., steps). In addition, the predictive accuracy on 13 of the 20 sampled nodes increases with a certain step larger than two.
This motivates the design of \emph{node-adaptive} aggregation functions, which determines the importance of a node's message at different ranges rather than fixing the same weights for all nodes. 

To this end, we propose the gating aggregator, which generates retainment scores that indicate how much the corresponding messages should be retained in the final combined message.

\begin{small}
\begin{equation}
\begin{aligned}
      & \mathbf{c_{msg}} \gets \sum_{\mathbf{m}_v^i\in M_v} w_i \mathbf{m}_v^i,\  w_i = \sigma(\mathbf{s} \mathbf{m}_v^i),\label{eq:gate}
\end{aligned}
\end{equation}
\end{small}
where $\mathbf{s}$ is a trainable vector to generate gating scores, and $\sigma$ is the sigmoid function. 
With the adaptive \texttt{message\_aggregator}, the model can balance the messages from the multi-scale neighborhoods for each node at the expense of training extra parameters. 

\begin{figure*}[t]
\vspace{-3mm}
	\centering
	\includegraphics[width=0.9\linewidth]{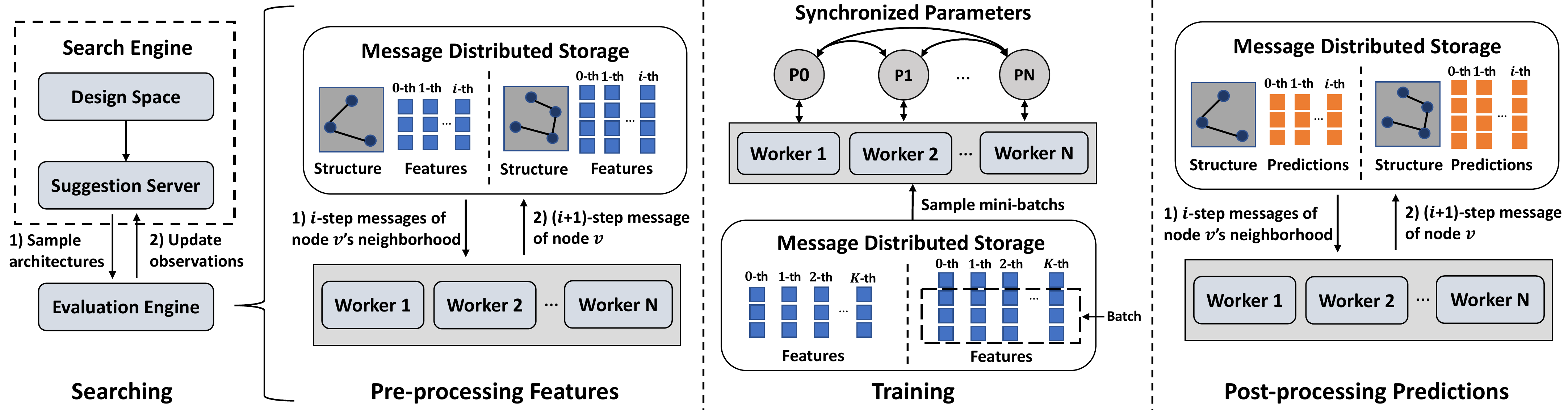}
	\vspace{-2mm}
	\caption{The workflow of \sys, which consists of the searching and evaluation engine.}
	\label{fig:process}
	\vspace{-4mm}
\end{figure*}

\subsection{SGAP Instances}
Recent scalable GNN models, such as SGC~\cite{wu2019simplifying}, SIGN~\cite{sign_icml_grl2020}, S$^2$GC~\cite{zhu2021simple} and GBP~\cite{chen2020scalable}, can be considered as specific instances in our defined design space, 
as shown in Table~\ref{sGNN}. We see that most current scalable GNNs ignore the post-processing stages, which can boost the model performance demonstrated by our experiments.

Besides, various scalable GNNs can be obtained by using different design choices under SGAP. For example, the current state-of-the-art scalable model GBP sets the $\texttt{graph\_aggregator}$ as Aug.NA and uses the weighted strategy in the $\texttt{message\_aggregator}$.
We decouple MLP training and information propagation in APPNP~\cite{klicpera2019predict} into two individual processes and get a new scalable model \sys-APPNP. \emph{To effectively explore the large design space, we implement an auto-search system engine below to automate the search procedure of scalable GNN architecture instead of manual design.}

\begin{table}[t]
\caption{Current scalable GNNs in our design space.}
\vspace{-2mm}
\centering
{
\noindent
\renewcommand{\multirowsetup}{\centering}
\resizebox{0.95\linewidth}{!}{
\begin{tabular}{c|c|cc|c}
\toprule
\multirow{2}{*}{\textbf{Models}}& \multicolumn{1}{c|}{\textbf{Pre-processing}} & 
\multicolumn{2}{c|}{\textbf{Model training}} & 
\multicolumn{1}{c}{\textbf{Post-processing}} \\
\cline{2-5}
 & \textbf{$GA_{pre}$}&\textbf{$MA$}&\textbf{$K_{trans}$}&\textbf{$GA_{post}$}\\
\hline
SGC & Aug.NA & None & 1 & /\\
SIGN & Optional & Concatenate & 1 & /\\
S$^2$GC & PPR & Mean & 1 & /\\
GBP & Aug.NA & Weighted & $\geq2$ & /\\
\sys-APPNP & / & / & $\geq2$ & PPR\\
\bottomrule
\end{tabular}}}
 \label{sGNN}
 \vspace{-3mm}
\end{table}

\section{\sys Engines}
Figure~\ref{fig:process} shows the overview of our proposed auto-search system to explore GNN designs under \sys abstraction. It consists of two engines: the \emph{search} engine and the \emph{evaluation} engine. The search engine includes the proposed designed search space for scalable GNNs and implements a suggestion server that is responsible for suggesting architectures to evaluate. The evaluation engine receives an instance and trains the corresponding architecture in a distributed fashion. 
An iteration of the searching process is as follows: 1) The suggestion server samples an architecture instance based on its built-in optimization algorithm and sends it to the evaluation engine; 2) The evaluation engine evaluates the architecture and updates the suggestion server with its observed performance.

\subsection{Search Engine}
While prior researches on scalable GNNs~\cite{sign_icml_grl2020,chen2020scalable} focus on optimizing the classification error, recent applications not only require high predictive performance but also \textit{low resource-consumption}, e.g. model size or inference time. 
In addition, there is typically an implicit trade-off between predictive performance and resource consumption. To this end, the suggestion server implements a \textit{multi-objective} search algorithm to tackle this trade-off. 

Concretely, we use the Bayesian optimization based on EHVI~\cite{emmerich2005single}, a widely-used algorithm that maximizes the predicted improvement of hypervolume indicator of Pareto-optimal points relative to a given reference point.  
The suggestion server then optimizes over the search space following a typical Bayesian optimization as 1) Based on the observation history, the server trains multiple surrogates, namely the Gaussian Process, to model the relationships between each architecture instance and its objective values; 2) The server randomly samples a number of new instances, and suggests the best one which maximizes the EHVI based on the predicted 
outputs of trained surrogates; 3) The server receives the results of the suggested instance and updates its observation history.

\subsection{Evaluation Engine}
Different from the sampling-training process of existing GNN systems 
(e.g., DistDGL~\cite{zheng2020distdgl}, NextDoor~\cite{jangda2021accelerating}, and FlexGraph~\cite{wang2021flexgraph}), the process of \sys evaluation engine is decoupled into pre-processing, training and post-processing as illustrated in Figure~\ref{fig:process}: \sys first \emph{pre-computes} the feature messages for each node over the graph, and then it \emph{combines} the messages and \emph{trains} the model parameters with parameter sever. Finally, \sys \emph{post-computes} the prediction messages for each node over the graph.
All the messages are partitioned and stored in a distributed storage system, and the stages can be implemented in a distributed fashion. Specifically, the engine consists of the following two components:

\para{Graph Data Aggregator.}
This component handles pre-processing and post-processing stages on data aggregation over graph structure.
The two stages share the same pipeline but take different messages as inputs (features for pre-processing and predictions for post-processing). We implement an efficient batch processing pipeline over distributed graph storage: The nodes are partitioned into batches, and the computation of each batch is implemented by workers in parallel with matrix multiplication. 
As shown in Figure~\ref{fig:process}, for each node in a batch, we firstly pull all the $i$-th step messages of its 1-hop neighbors from the message distributed storage and then compute the $(i+1)$-th step messages of the batch in parallel. Next, We push these aggregated messages back for reuse in the calculation of the $(i+2)$-th step messages. 
In our implementation, we treat GPUs as workers for fast processing, and the graph data are partitioned and stored on host memory across machines. 
Given the parallel message computation, our implementation could scale to large graphs and significantly reduce the runtime.

\para{Neural Architecture Trainer.}
This component handles the training of neural networks. We optimize the parameters of each architecture with distributed SGD.
The model parameters are stored on a parameter server, and multiple workers (GPUs) process the data in parallel. 
We adopt asynchronous training to avoid the communication overhead between workers. 
Each worker fetches the most up-to-date parameters and computes the gradients
for a mini-batch of data, independent of the other workers.

\section{Experiments}
\subsection{Experimental Settings}
\para{Datasets.}
We conduct the experiments on three citation networks (Citeseer, Cora, and PubMed) in~\cite{kipf2017semi}, two social networks (Flickr and Reddit) in~\cite{zeng2020graphsaint}, four co-authorship graphs (Amazon and Coauthor) in~\cite{pei2020geom}, the co-purchasing network (ogbn-products) in~\cite{hu2020open} and one short-form video recommendation graph (Industry) from our industrial cooperative enterprise.
Table~\ref{Datasets} in Appendix~\ref{appendix:dataset} provides the overview of the used graph datasets.

\para{Parameters and Environment.}
To eliminate random factors, we run each method 20 times and report the mean and variance of the performance. 
More details for experimental setups and reproduction are provided in Appendix~\ref{appendix:setup} and ~\ref{appendix:reproduction}.

\para{Baselines.}
In the transductive settings, we compare the searched scalable GNNs with GCN~\cite{kipf2017semi}, GAT~\cite{velivckovic2018graph}, JK-Net~\cite{xu2018representation}, Res-GCN~\cite{kipf2017semi}, APPNP~\cite{klicpera2019predict}, AP-GCN~\cite{spinelli2020adaptive}, SGC~\cite{wu2019simplifying}, SIGN~\cite{sign_icml_grl2020}, S$^2$GC~\cite{zhu2021simple} and GBP~\cite{chen2020scalable},
which are SOTA models of different message passing types. 
In the inductive settings, the compared baselines are GraphSAGE~\cite{hamilton2017inductive}, FastGCN~\cite{chen2018fastgcn}, ClusterGCN~\cite{chiang2019cluster} and GraphSAINT~\cite{zeng2020graphsaint}. More descriptions about the baselines are provided in Appendix~\ref{appendix:baseline}.

In the following, we first analyze the superiority of representative instances searched by \sys. Then we evaluate the transferability, training efficiency, and model scalability of \sys representatives compared with competitive state-of-the-art baselines.

\begin{figure*}[htbp]
\vspace{-2mm}
\centering
\subfigure[Stand-alone on Reddit]{
    \begin{minipage}[t]{0.24\linewidth}
    \centering
    \includegraphics[width=0.9\linewidth]{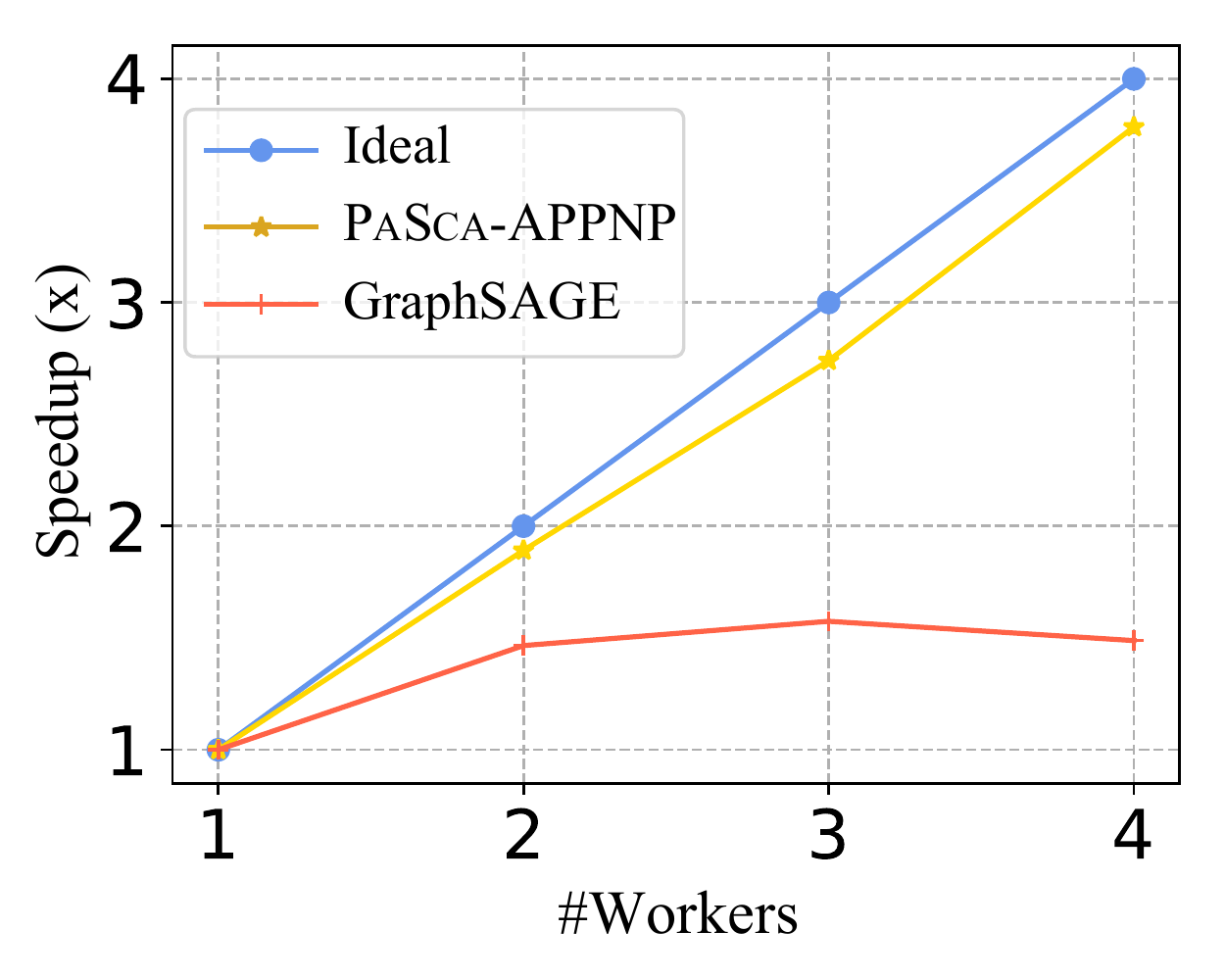}
    \end{minipage}%
    }
\subfigure[Distributed on Reddit]{
    \begin{minipage}[t]{0.24\linewidth}
    \centering
    \includegraphics[width=0.9\linewidth]{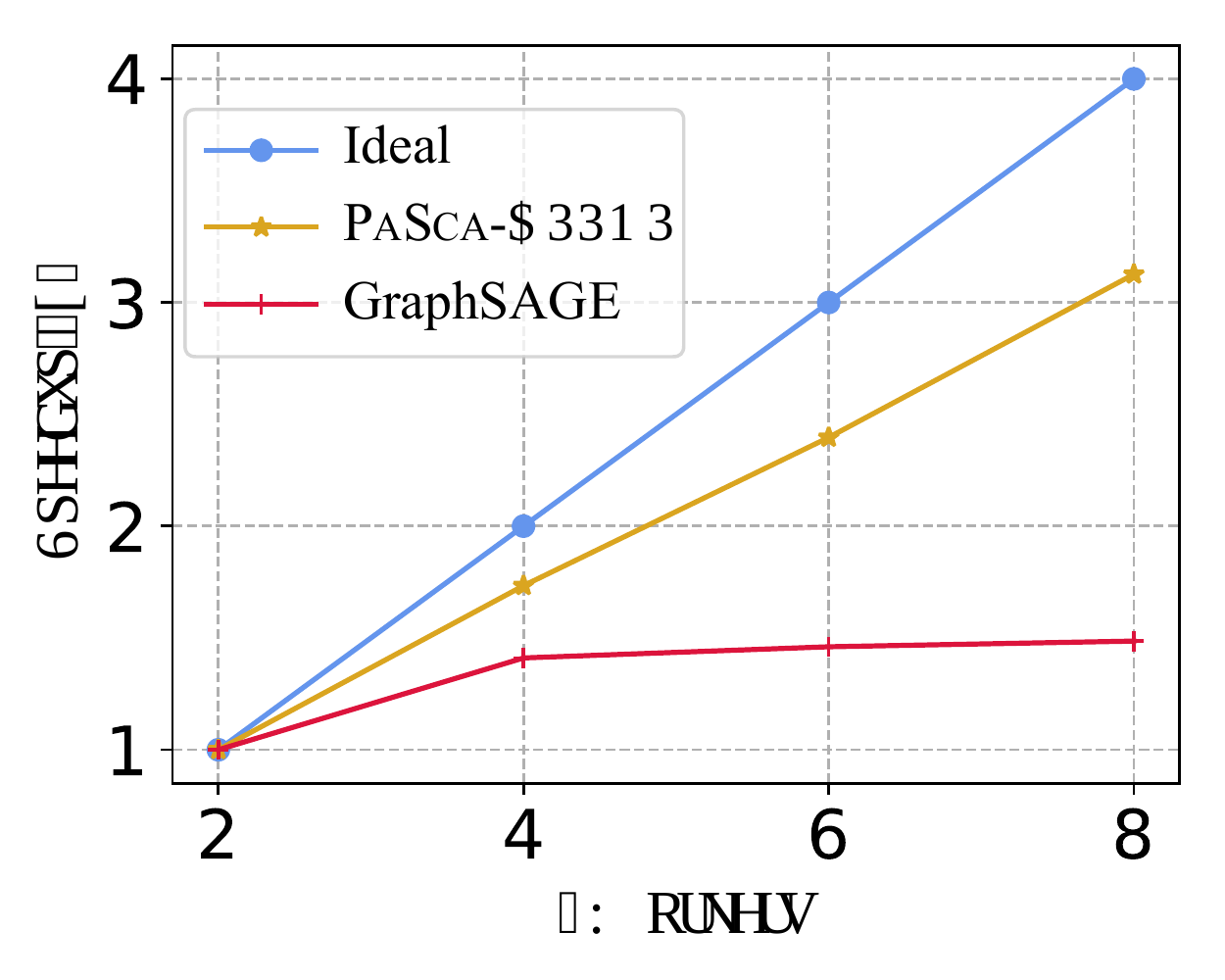}
    \end{minipage}%
    }%
\subfigure[Stand-alone on ogbn-product]{
    \begin{minipage}[t]{0.24\linewidth}
    \centering
    \includegraphics[width=0.9\linewidth]{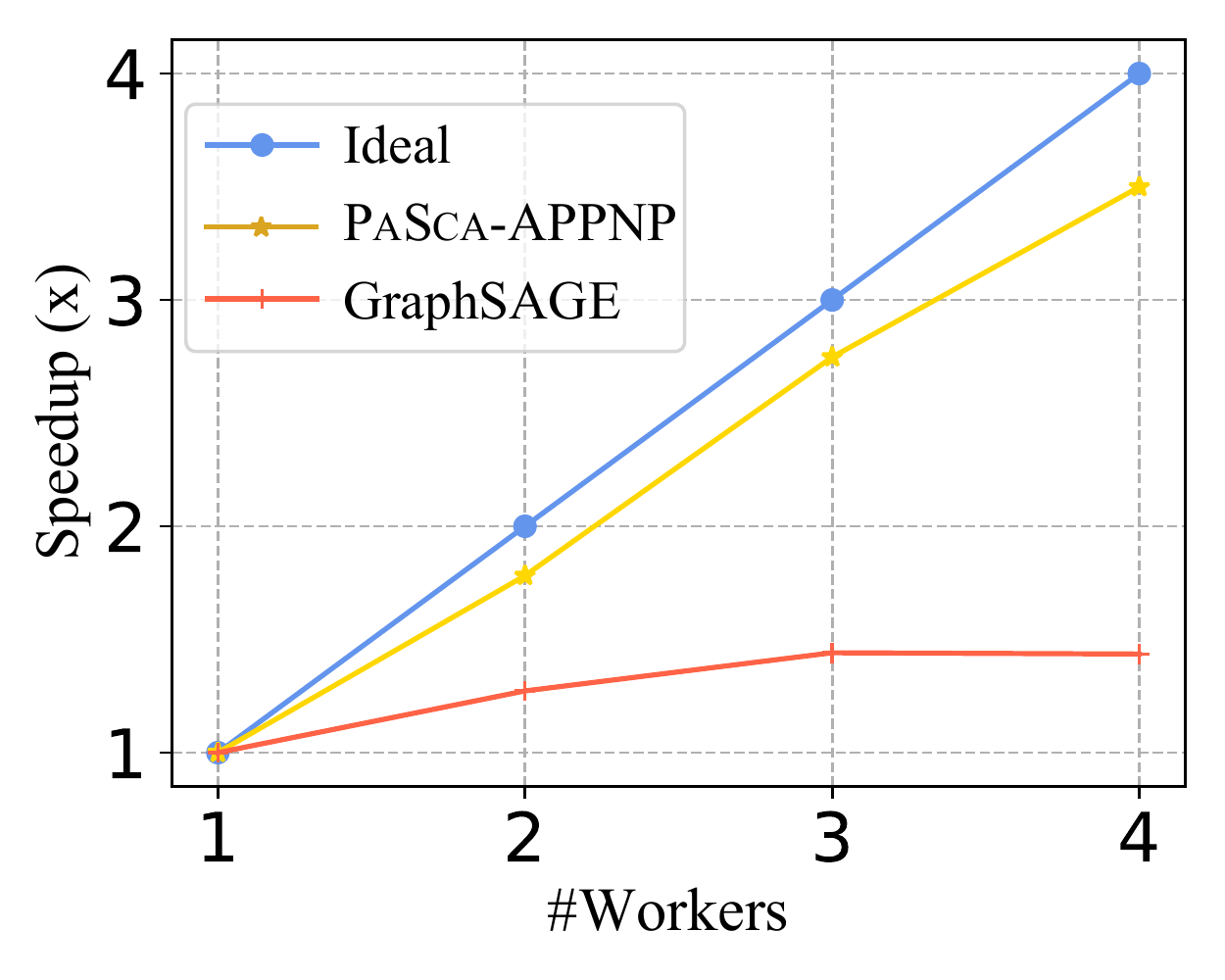}
    \end{minipage}
    }%
\subfigure[Distributed on ogbn-product]{
    \begin{minipage}[t]{0.24\linewidth}
    \centering
    \includegraphics[width=0.9\linewidth]{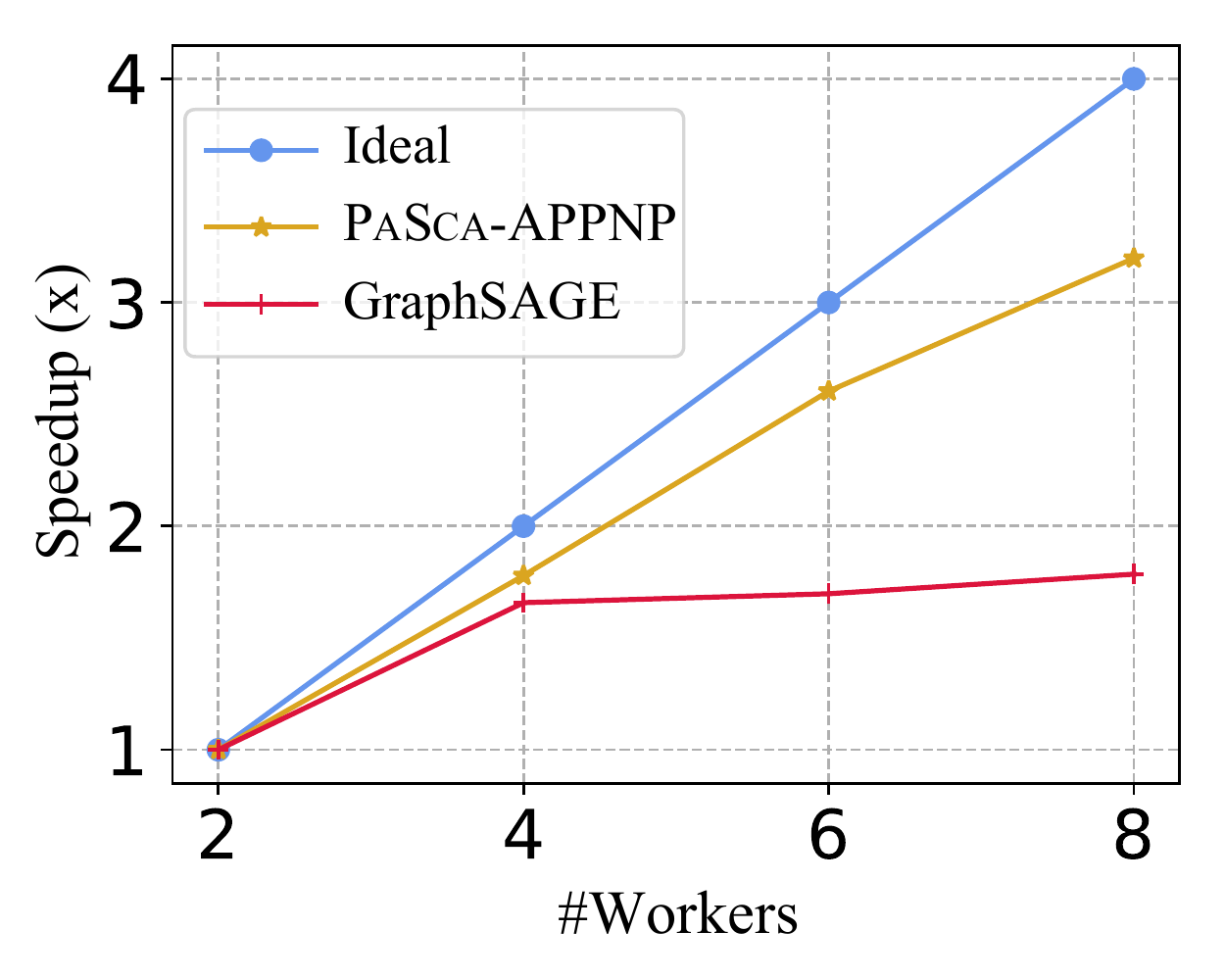}
    \end{minipage}
    }%
\vspace{-2mm}
\centering
\caption{Scalability comparison on Reddit and ogbn-product datasets. The stand-alone scenario means the graph has only one partition stored on a multi-GPU server, whereas the distributed scenario means the graph is partitioned and stored on multi-servers. In the distributed scenario, we run two workers per machine.}
\label{scalability}
\vspace{-3mm}
\end{figure*}

\begin{table}[tpb]
\caption{Scalable GNNs found by \sys.}
\vspace{-3mm}
\centering
{
\noindent
\renewcommand{\multirowsetup}{\centering}
\resizebox{1\linewidth}{!}{
\begin{tabular}{c|ccc|c|cc}
\toprule
\multirow{2}{*}{\textbf{Models}}& \multicolumn{3}{c|}{\textbf{Pre-processing}} & 
\multicolumn{1}{c|}{\textbf{Model training}} & 
\multicolumn{2}{c}{\textbf{Post-processing}} \\
\cline{2-7}
 & \textbf{$GA_{pre}$}&\textbf{$MA$}&\textbf{$K_{pre}$}&\textbf{$K_{trans}$}&\textbf{$GA_{post}$}&\textbf{$K_{post}$}\\
\hline
\sys-V1&PPR($\alpha$ = 0.1)&Weighted & 3 & 2 & / & /\\
\sys-V2&Aug.NA&Adaptive & 6 & 2 & / & /\\
\sys-V3&Aug.NA&Adaptive & 6 & 3 &PPR ($\alpha$ = 0.3) & 4\\
\bottomrule
\end{tabular}}}
 \label{pasca_sgnn}
\vspace{-4mm}
\end{table}

\begin{figure}[tpb]
    \centering
    \vspace{-5mm}
    \includegraphics[width=0.8\linewidth]{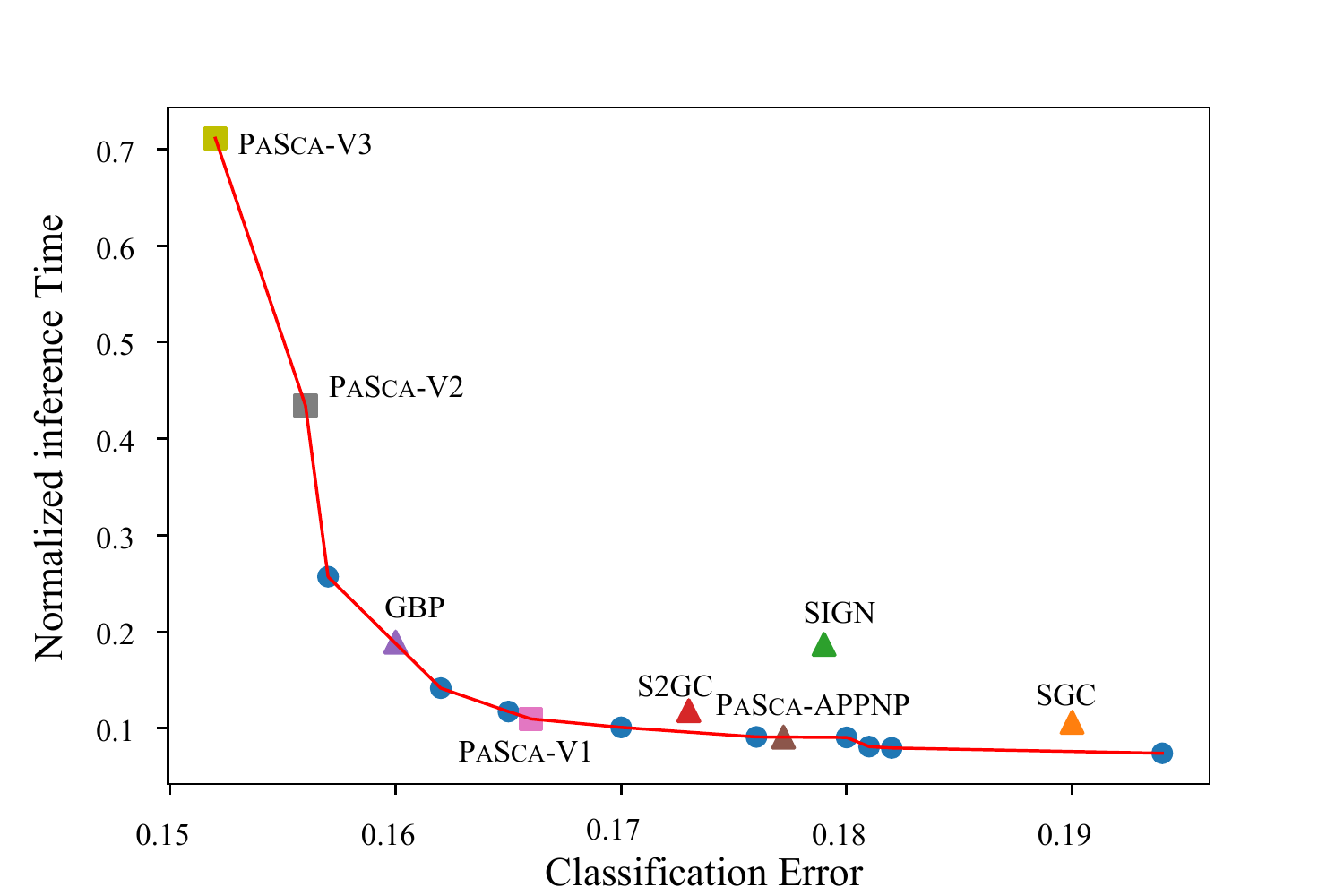}
    \vspace{-3mm}
    \caption{Pareto Front found on Cora.}
    \vspace{-3mm}
    \label{pareto}
\end{figure}

\subsection{Searched Representatives}
We apply the multi-objective optimization targeting at classification error and inference time on Cora. Figure~\ref{pareto} demonstrates the Pareto Front found by \sys with a budget of 2000 evaluations, together with the results of several manually designed scalable GNNs.
The inference time has been normalized based on instances with the minimum and maximum inference time in our design space. Interestingly, we observe that GBP and \sys-APPNP, our extended variant of APPNP (see Table~\ref{sGNN}), falls on the Pareto Front, which indicates the superior design of the ``Weighted'' \texttt{message\_aggregator} and ``PPR'' \texttt{graph\_aggregator}. 
We also choose other three instances from the Pareto Front as \sys-V1 to V3 with different accuracy-efficiency requirements as searched representatives of SGAP for the following evaluations.
The corresponding parameters of each architecture are shown in Table~\ref{pasca_sgnn}.
Among the three architectures, \sys-V1 Pareto-dominates the other baselines except GBP, and \sys-V3 is the architecture with the best predictive performance found by the search engine.

\subsection{Training Scalability}
The main characteristic of our proposed design space is that the architectures sampled from the space, namely \sys-SGAP, share high scalability upon workers.
To examine the scalability of \sys-SGAP, we choose \sys-APPNP as a representative and compare it with GraphSAGE, a widely-used method in industry on two large-scale datasets. We train GraphSAGE with DGL and \sys-APPNP with the evaluation engine of \sys, respectively.  
We train both methods in stand-alone and distributed scenarios and then measure their corresponding speedups.
The batch size is 8192 for Reddit and 16384 for ogbn-product, and the speedup is calculated by runtime per epoch relative to that of one worker in the stand-alone scenario and two workers in the distributed scenario.
Without considering extra cost, the speedup will increase linearly in an ideal condition.

The corresponding results are shown in Figure~\ref{scalability}.
Since GraphSAGE requires aggregating the neighborhood nodes during training, GraphSAGE trained with DGL meets the I/O bottleneck when transmitting a large number of required neural messages.
Thus, the speedup of GraphSAGE training increases slowly as the number of workers grows, which is less than 2$\times$ even with four workers in the stand-alone scenario and eight workers in the distributed scenario.
Recall that the only communication cost of \sys-SGAP is to synchronize parameters among different workers, which is essential to all distributed training methods.
As a result, \sys-SGAP trained with the evaluation engine scales up close to the ideal circumstance in both scenarios, indicating the superiority of \sys.

\begin{figure}[tpb]
    \centering
    \includegraphics[width=0.75\linewidth]{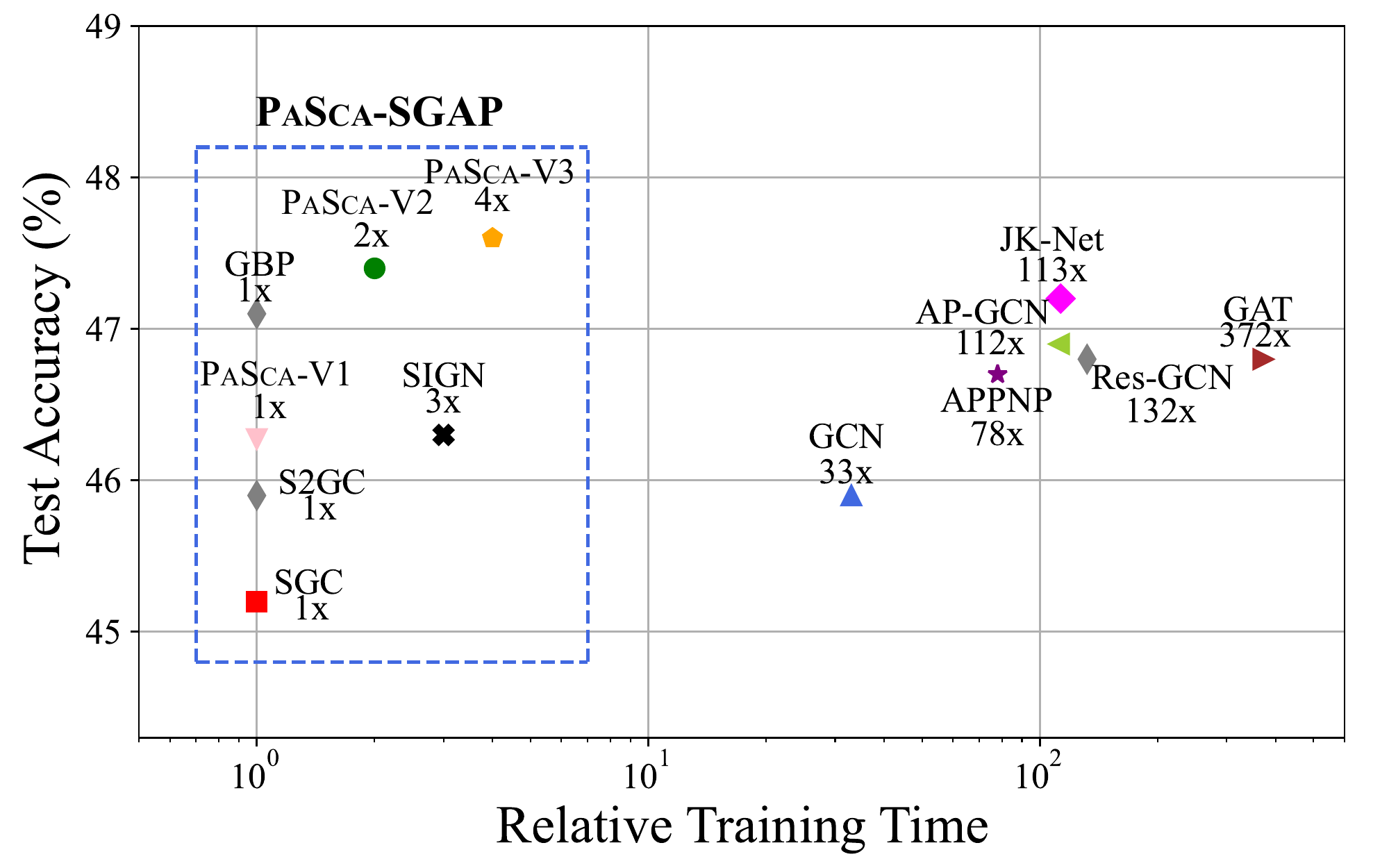}
    \vspace{-3mm}
    \caption{Test accuracy over training time on Industry.}
    \vspace{-5mm}
    \label{acc-efficiency}
\end{figure}

\begin{table*}[t]
\caption{Test accuracy (\%) in transductive settings. ``NMP'' and ``DNMP'' refer to architectures following NMP and DNMP paradigm. ``SGAP'' refers to architectures following the scalable graph archtecture paradigm proposed in Section~\ref{paradigm}.} 
\vspace{-2mm}
\centering
{
\noindent
\renewcommand{\multirowsetup}{\centering}
\resizebox{0.9\linewidth}{!}{
\begin{tabular}{cccccccccc}
\toprule
\textbf{Type}&\textbf{Models}&\textbf{Cora}& \textbf{Citeseer}&\textbf{PubMed}&{\textbf{\makecell{Amazon \\Computer}}} & 
{\textbf{\makecell{Amazon \\Photo}}} & 
{\textbf{\makecell{Coauthor \\CS}}}&
{\textbf{\makecell{Coauthor \\Physics}}}&
{\textbf{\makecell{Industry}}}\\
\midrule
\multirowcell{4}{NMP} & 
GCN& 81.8$\pm$0.5 & 70.8$\pm$0.5 &79.3$\pm$0.7&82.4$\pm$0.4 & 91.2$\pm$0.6 & 90.7$\pm$0.2 & 92.7$\pm$1.1& 45.9$\pm$0.4 \\
&GAT& 83.0$\pm$0.7 & 72.5$\pm$0.7 &79.0$\pm$0.3&80.1$\pm$0.6 & 90.8$\pm$1.0 & 87.4$\pm$0.2 & 90.2$\pm$1.4& 46.8$\pm$0.7 \\
&JK-Net& 81.8$\pm$0.5  & 70.7$\pm$0.7 & 78.8$\pm$0.7 & 82.0$\pm$0.6 & 91.9$\pm$0.7 & 89.5$\pm$0.6 & 92.5$\pm$0.4& 47.2$\pm$0.3 \\
&ResGCN& 82.2$\pm$0.6 & 70.8$\pm$0.7 & 78.3$\pm$0.6& 81.1$\pm$0.7 & 91.3$\pm$0.9 & 87.9$\pm$0.6 & 92.2$\pm$1.5& 46.8$\pm$0.5 \\
\midrule
\multirowcell{2}{DNMP} &
APPNP& 83.3$\pm$0.5 & 71.8$\pm$0.5 & 80.1$\pm$0.2&81.7$\pm$0.3&91.4$\pm$0.3&92.1$\pm$0.4&92.8$\pm$0.9 &46.7$\pm$0.6\\
&AP-GCN& 83.4$\pm$0.3& 71.3$\pm$0.5& 79.7$\pm$0.3&83.7$\pm$0.6& 92.1$\pm$0.3& 91.6$\pm$0.7& 93.1$\pm$0.9&46.9$\pm$0.7\\
\midrule
\multirowcell{7}{SGAP}&
SGC & 81.0$\pm$0.2 & 71.3$\pm$0.5 & 78.9$\pm$0.5&82.2$\pm$0.9&91.6$\pm$0.7&90.3$\pm$0.5& 91.7$\pm$1.1 &45.2$\pm$0.3\\
&SIGN& 82.1$\pm$0.3 & 72.4$\pm$0.8 &79.5$\pm$0.5&83.1$\pm$0.8&91.7$\pm$0.7&91.9$\pm$0.3& 92.8$\pm$0.8  &46.3$\pm$0.5\\
&S$^2$GC & 82.7$\pm$0.3 & 73.0$\pm$0.2 & 79.9$\pm$0.3&83.1$\pm$0.7&91.6$\pm$0.6&91.6$\pm$0.6& 93.1$\pm$0.8 &45.9$\pm$0.4\\
&GBP& 83.9$\pm$0.7 & 72.9$\pm$0.5 &80.6$\pm$0.4&83.5$\pm$0.8&92.1$\pm$0.8&92.3$\pm$0.4& 93.3$\pm$0.7  &47.1$\pm$0.6\\
&\sys-V1&83.4$\pm$0.5&72.2$\pm$0.5&80.5$\pm$0.4&83.7$\pm$0.7&92.1$\pm$0.7&91.9$\pm$0.3& 93.2$\pm$0.6&46.3$\pm$0.4\\
&\sys-V2&84.4$\pm$0.3&73.1$\pm$0.3&80.7$\pm$0.7&84.1$\pm$0.7&92.4$\pm$0.7&92.6$\pm$0.4& 93.6$\pm$0.8&47.4$\pm$0.6\\
&\sys-V3& \textbf{84.6$\pm$0.6} & \textbf{73.4$\pm$0.5} &\textbf{80.8$\pm$0.6}&\textbf{84.8$\pm$0.7}&\textbf{92.7$\pm$0.8}  &\textbf{92.8$\pm$0.5}& \textbf{93.8$\pm$0.9}&\textbf{47.6$\pm$0.3}\\ 
\bottomrule
\end{tabular}}}
\label{Node1}
\vspace{-2mm}
\end{table*}

\begin{table}[th]
\caption{Test accuracy (\%) in inductive settings.}
\vspace{-2mm}
\centering
{
\noindent
\renewcommand{\multirowsetup}{\centering}
\resizebox{0.6\linewidth}{!}{
\begin{tabular}{cccc}
\toprule
\textbf{Models}& \textbf{Flickr}&\textbf{Reddit}\\
\midrule
GraphSAGE & 50.1$\pm$1.3 & 95.4$\pm$0.0 \\
FastGCN & 50.4$\pm$0.1 & 93.7$\pm$0.0 \\
ClusterGCN & 48.1$\pm$0.5 & 95.7$\pm$0.0 \\
GraphSAINT & 51.1$\pm$0.1 & 96.6$\pm$0.1\\
\midrule
\sys-V1&51.2$\pm$0.3&95.8$\pm$0.1\\
\sys-V2&51.8$\pm$0.3&96.3$\pm$0.0\\
\sys-V3& \textbf{52.1$\pm$0.2}  & \textbf{96.7$\pm$0.1}  \\
\bottomrule
\end{tabular}}}
\label{Inductive}
\vspace{-4mm}
\end{table}

\subsection{Performance-Efficiency Analysis}
To test the transferability and training efficiency of \sys models, we further evaluate them on more datasets compared with competitive baselines.
The results are summarized in Table~\ref{Node1} and \ref{Inductive}. 

We observe that \sys models obtain quite competitive performance in both transductive and inductive settings. 
In transductive settings, our simplified variant \sys-V1 also achieves the best performance among Non-SGAP baselines on most datasets, which shows the superiority of SGAP design.
In addition, \sys-V2 and V3 outperform the best baseline GBP by a margin of 0.1\%$\sim$0.6\% and 0.2\%$\sim$1.3\% on each dataset. We attribute this improvement to the application of the adaptive \texttt{message\_aggregator}. 
In inductive settings, Table \ref{Inductive} shows that \sys-V3 outperforms the best baseline GraphSAINT by a margin of 1.1\% on Flickr and 0.1\% on Reddit.

We also evaluate the training efficiency of each method in the real production environment.
Figure \ref{acc-efficiency} illustrates the performance over training time on Industry. 
In particular, we pre-compute the feature messages of each scalable method, and the training time takes into account the pre-computation time. 
We observe that NMP architectures require at least a magnitude of training time than \sys-SGAP. 
Among considered baselines, \sys-V3 achieves the best performance with 4$\times$ training time compared with GBP and \sys-V1. 
Note that, though \sys-V1 requires the same training time as GBP, its inference time is less than GBP. 
Therefore, we recommend choosing \sys-V1 to V3, along with GBP, according to different requirements of predictive performance, training efficiency, and inference time.

\subsection{Model Scalability}
\begin{figure}[tpb]
    \centering
    \subfigure{
    \scalebox{0.4}{
    \includegraphics[width=0.98\linewidth]{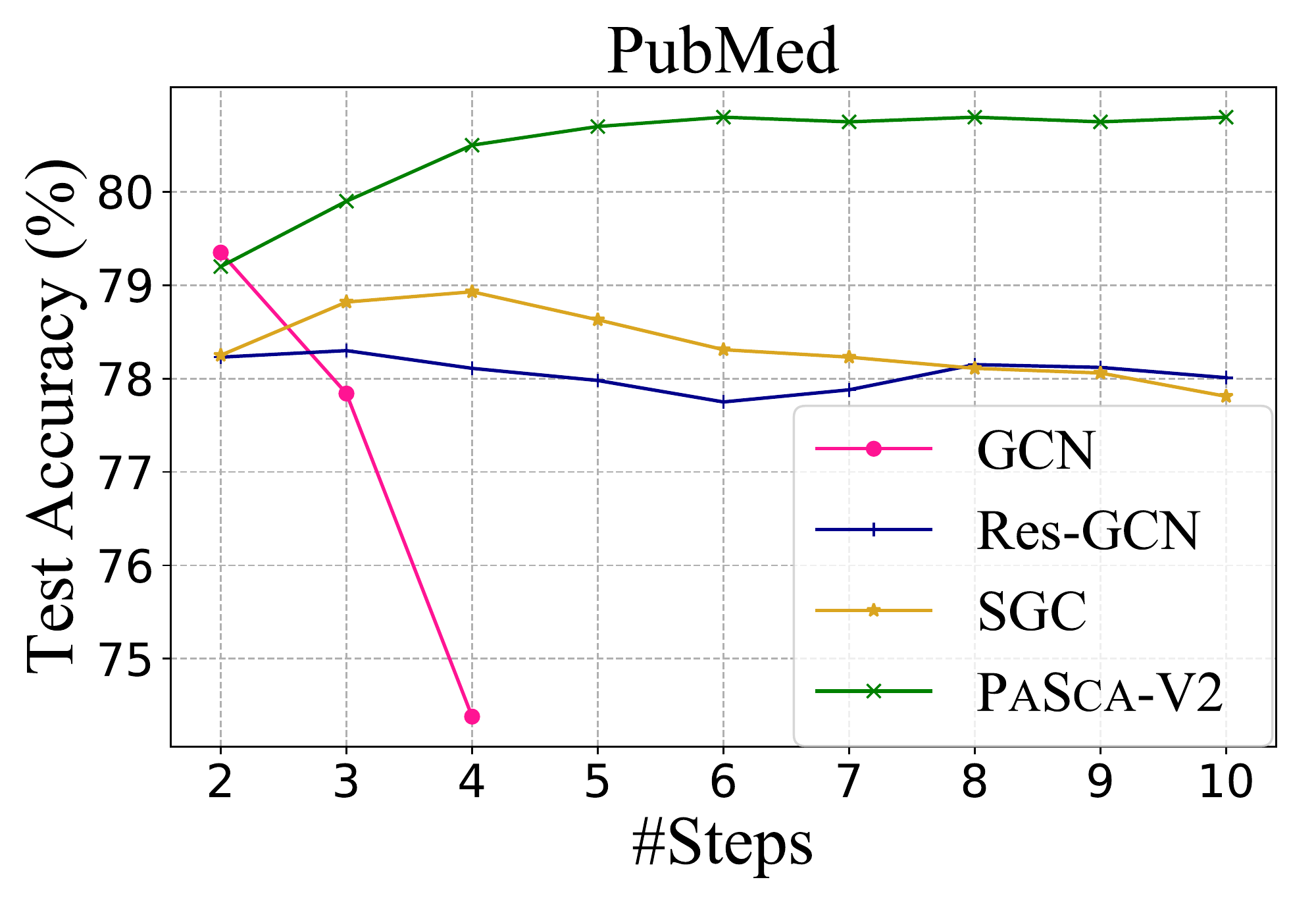}}
    }
    \subfigure{
    \scalebox{0.56}{
    \includegraphics[width=0.98\linewidth]{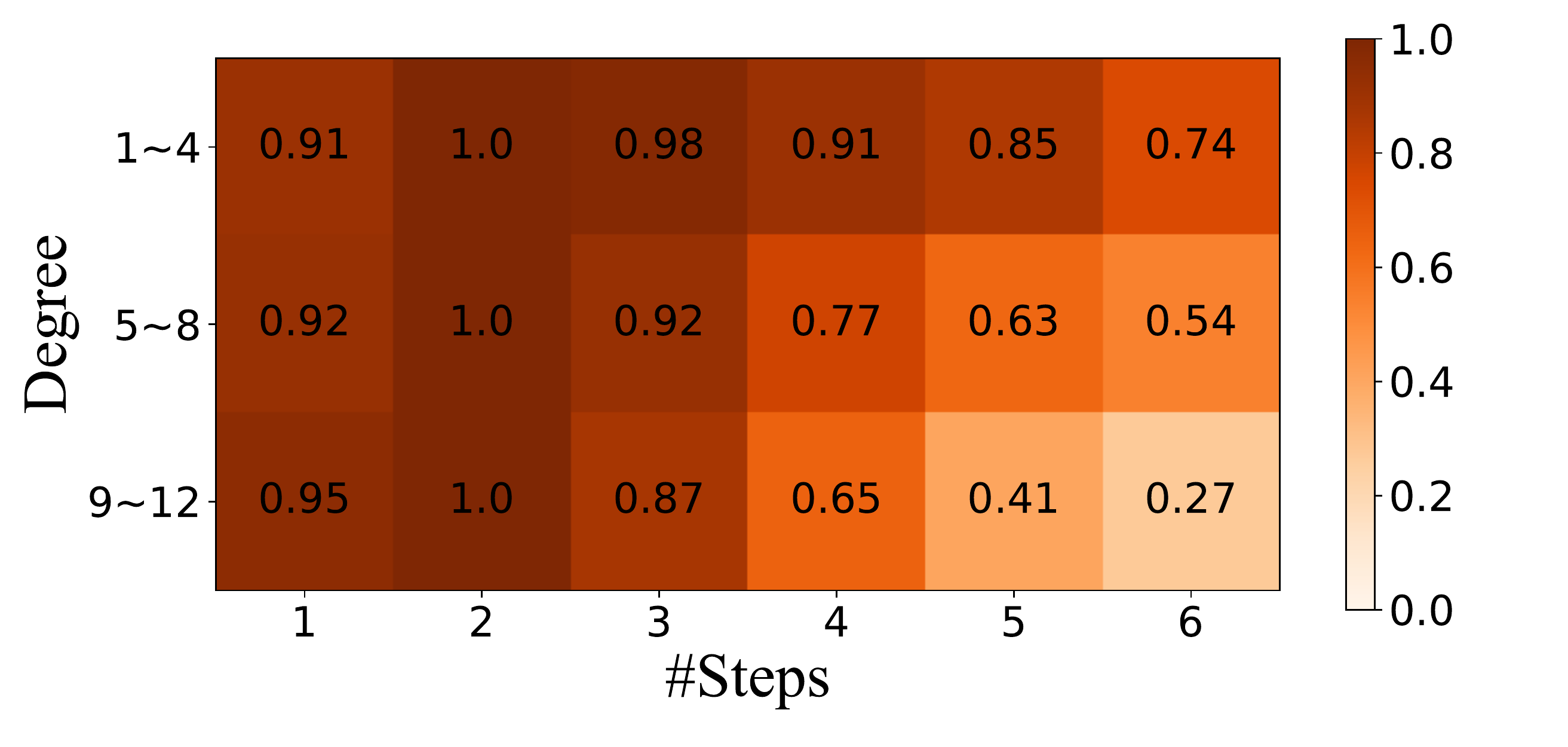}}
    }
    \vspace{-5mm}
    \caption{Left: Test accuracy of different models along with the increased aggregation steps on PubMed. Right: \sys-V2's average gating weights of graph messages of different steps on 60 randomly selected nodes from PubMed.}
    \label{interpretability}
    \vspace{-2mm}
\end{figure}

We observe that both \sys-V2 and V3 found by the search engine contain the ``Adaptive'' \texttt{message\_aggregator}. In this subsection, we aim to explain the advantage of adaptive \texttt{message\_aggregator} in the perspective of model scalability on message passing steps.
We plot the changes of model performance along with the message passing steps in the left subfigure of Figure~\ref{interpretability}. For a fair comparison, we use \sys-V2 which does not include post-processing. The vanilla GCN gets the best results with two aggregation steps, but its performance drops rapidly along with the increased steps due to the over-smoothing issue.
Both Res-GCN and SGC show better performance than GCN with larger aggregation steps. 
Take Res-GCN as an example, it carries information from the previous step by introducing the residual connections and thus alleviates this problem.
However, these two methods cannot benefit from deep GNN architecture since they are unable to balance the needs of preserving locality (i.e., staying close to the root node to avoid over-smoothing) and leveraging the information from a large neighborhood.
In contrast, \sys-V2 achieves consistent improvement and remains non-decreasing across steps, which indicates that \sys can scales to large depth. 
The reason is that the adaptive \texttt{message\_aggregator} in \sys-V2 is able to adaptively and effectively combine multi-scale neighborhood messages for each node.

To demonstrate this, the right subfigure of Figure \ref{interpretability} shows \sys-V2's average gating weights of feature messages according to the number of steps and degrees of input nodes, where the maximum step is 6. In this experiment, we randomly select 20 nodes for each degree range (1-4, 5-8, 9-12) and plot the relative weight based on the maximum value. 
We get two observations from the heat map: 1) The 1-step and 2-step graph messages are always of great importance, which shows that the adaptive \texttt{message\_aggregator} captures the local information as those widely used 2-layer GNNs do; 2) The weights of graph messages with larger steps drop faster as the degree grows, which indicates that the attention-based aggregator could prevent high-degree nodes from including excessive irrelevant nodes which lead to over-smoothing. From the two observations, we conclude that the adaptive \texttt{message\_aggregator} can identify the different message-passing demands of nodes and explicitly weight each graph message.

\section{Conclusion}
In this paper, we proposed \sys, a new auto-search system that offers a principled approach to systemically construct and explore the design space for scalable GNNs, rather than studying individual designs.
Experiments on ten real-world benchmarks demonstrate that the representative instances searched by \sys outperform SOTA GNNs in terms of performance, efficiency, and scalability. \sys can help researchers understand design choices when developing new scalable GNN models, and serve as a system to support extensive exploration over the design space for scalable GNNs.

\section*{Acknowledgments}
This work is supported by NSFC (No. 61832001, 61972004), Beijing Academy of Artificial Intelligence (BAAI), PKU-Baidu Fund 2019BD006, and PKU-Tencent Joint Research Lab. 
Zhi Yang and Bin Cui are the corresponding authors.

\clearpage

\bibliographystyle{ACM-Reference-Format}
\balance
\bibliography{my-reference}

\clearpage
\appendix
\section{Appendix}

\subsection{Dataset description}
\label{appendix:dataset}
\begin{table*}[t]
\small
\centering
\caption{Overview of the Graph Datasets} \label{Datasets}
\begin{tabular}{ccccccccc}
\toprule
\textbf{Dataset}&\textbf{\#Nodes}& \textbf{\#Features}&\textbf{\#Edges}&\textbf{\#Classes}&\textbf{\#Train/Val/Test}&\textbf{Task type}&\textbf{Description}\\
\midrule
Cora& 2,708 & 1,433 &5,429&7& 140/500/1000 & Transductive&citation network\\
Citeseer& 3,327 & 3,703&4,732&6& 120/500/1000 & Transductive&citation network\\
Pubmed& 19,717 & 500 &44,338&3& 60/500/1000 & Transductive&citation network\\
\midrule
Amazon Computer& 13,381  & 767& 245,778 & 10 &200/300/12881&Transductive&co-purchase graph\\
Amazon Photo &7,487  & 745& 119,043 & 8 & 160/240/7,087&Transductive&co-purchase graph\\
ogbn-products&2,449,029&100&61,859,140&47&195922/489811/204126&Transductive&co-purchase network\\
Coauthor CS& 18,333  & 6,805 & 81,894 & 15& 300/450/17,583&Transductive&co-authorship graph \\
Coauthor Physics& 34,493 & 8,415 & 247,962 & 5& 100/150/34,243&Transductive&co-authorship graph \\
\midrule
Flickr& 89,250 & 500 & 899,756 & 7 &  44,625/22,312/22,312 & Inductive &image network\\
Reddit& 232,965 & 602 & 11,606,919 & 41 &  155,310/23,297/54,358 & Inductive&social network \\
\midrule
Industry & 1,000,000 & 64 & 1,434,382 & 253 & 5,000/10,000/30,000&Transductive&user-video graph\\
\bottomrule
\end{tabular}
\end{table*}

\textbf{Cora}, \textbf{Citeseer}, and \textbf{Pubmed}\footnote{https://github.com/tkipf/gcn/tree/master/gcn/data} are three well-known citation network datasets, and we follow the same training/validation/test split as GCN~\cite{kipf2017semi}.

\noindent\textbf{Reddit} is a social network modeling the community structure of Reddit posts. This dataset is often used for inductive training, and the training/validation/test split is coherent with that of GraphSAGE~\cite{hamilton2017inductive}.

\noindent\textbf{Flickr} originates from NUS-wide~\footnote{http://lms.comp.nus.edu.sg/research/NUS-WIDE.html} and contains different types of images based on the descriptions and common properties of online images. We use a public version of Reddit and Flickr provided by GraphSAINT\footnote{https://github.com/GraphSAINT/GraphSAINT}.

\noindent\textbf{Amazon Computers} and \textbf{Amazon Photo} are segments of the Amazon co-purchase graph~\cite{shchur2018pitfalls}, where nodes represent goods, edges indicate that two goods are frequently bought together, node features are bag-of-words encoded product reviews, and class labels are given by the product category.

\noindent\textbf{Coauthor CS} and \textbf{Coauthor Physics} are co-authorship graph based on the Microsoft Academic Graph from the KDD Cup 2016 challenge\footnote{https://kddcup2016.azurewebsites.net/}. 
Here, nodes are authors, that are connected by an edge if they co-authored a paper; node features represent paper keywords for each author's papers, and class labels indicate the most active fields of study for each author. We use a pre-divided version of these datasets through the Deep Graph Library (DGL)\footnote{https://docs.dgl.ai/en/0.4.x/api/python/data.html\#coauthor-dataset}.

\noindent\textbf{ogbn-products} is an unweighted graph representing an Amazon product co-purchase network.
Each node represents a product sold on Amazon, and edges between two products indicate that the products are purchased together.
We use the public data split for this dataset as in Open Graph Benchmark\footnote{https://github.com/snap-stanford/ogb}.

\noindent\textbf{Industry} is a user-video graph collected from a real-world mobile application from our industry partner. 
 We sampled 1,000,000 users and videos from the app, and treat these items as nodes. The edges in the generated bipartite graph represent that the user clicks the short videos. 
 Each user has 64 features, and the target is to category these short videos into 253 different classes. 
 
 \subsection{Decoupled Neural Message Passing} 
 \label{DNMP}
Note that the aggregate and update operations are inherently intertwined in Equation~\eqref{eq:mp}, i.e., each aggregate operation requires a neural layer to update the node's hidden state in order to generate a new message for the next step. Recently, some researches show that such entanglement could compromise performance on a range of benchmark tasks~\cite{wu2019simplifying, sign_icml_grl2020, zhang2021graph}, and suggest separating GCN from the aggregation scheme.
We reformulate these models into a single Decoupled Neural Message Passing (DNMP) framework: Neural prediction messages are first generated (with update function) for each node
utilizing only that node’s own features, and then aggregated using aggregate function.
\begin{equation}
\begin{aligned}
& \mathbf{h}^{0}_v \gets \texttt{update}(\mathbf{x}_v),\  \mathbf{h}^t_v \gets \texttt{aggregate}\left(\left\{\mathbf{h}^{t-1}_u|{u \in \mathcal{N}_v}\right\}\right).
\end{aligned}
\label{eq:dmp}
\end{equation}
where $x_v$ is the input feature of node $v$. 
Existing methods, such as PPNP~\cite{klicpera2019predict}, APPNP~\cite{klicpera2019predict}, AP-GCN~\cite{spinelli2020adaptive} and etc., follows this decoupled MP. Taking APPNP as an example:
\begin{equation}
\begin{aligned}
&\texttt{APPNP-update}(\mathbf{x_v})=\sigma(W\mathbf{x}_v),\\
&\texttt{APPNP-aggregate}\left(\left\{\mathbf{h}^{t-1}_u|{u \in \mathcal{N}_v}\right\}\right)=\alpha \mathbf{h}_v^0+(1-\alpha)\sum_{u \in \mathcal{N}_v}\frac{\mathbf{h}_u^{t-1}}{\sqrt{\tilde{d}_v\tilde{d}_u}}, 
\end{aligned}\nonumber
\end{equation}
where \texttt{aggregate} function adopts personalized PageRank with the restart probability $\alpha \in \left(0,1 \right]$ controlling the locality.

\subsection{More details about the compared baselines}
\label{appendix:baseline}
The main characteristic of all baselines are listed as follows:
\begin{itemize}
    \item \textbf{GCN}~\cite{kipf2017semi} produces node embedding vectors by truncating the Chebyshev polynomial to the first-order neighborhoods.
    
    \item \textbf{ResGCN}~\cite{kipf2017semi} adopts the residual connections between hidden layers to facilitate the training of deeper models by enabling the model to carry over information from the previous layer's input.

    \item \textbf{JK-Net}~\cite{xu2018representation} proposes a new aggregation scheme for node representation learning that can adapt neighborhood ranges to nodes individually.
    
    \item \textbf{APPNP}~\cite{klicpera2019predict} uses the relationship between GCN and PageRank to derive an improved propagation scheme based on personalized PageRank.
    
    \item \textbf{AP-GCN}~\cite{spinelli2020adaptive} is a variation of GCN wherein each node selects automatically the number of propagation steps performed across the graph.
    
    \item \textbf{SGC}~\cite{wu2019simplifying} reduces the excess complexity of GCN through successively removing non-linearities and collapsing weight matrices between consecutive layers.
    
    \item \textbf{SIGN}~\cite{sign_icml_grl2020} is a sampling-free Graph Neural Network model that is able to easily scale to gigantic graphs while retaining enough expressive power.
    
    \item \textbf{GraphSAGE}~\cite{hamilton2017inductive} is an inductive framework that leverages node attribute information to efficiently generate representations on previously unseen data.
    
     \item \textbf{GAT}~\cite{liao2018graph} leverages masked self-attentional layers to address the shortcomings of prior GNNs based on graph convolutions or their approximations, and enables specifying different weights to different nodes in a neighborhood.
    
    \item \textbf{S$^2$GC} ~\cite{zhu2021simple}: S$^2$GC uses a modified Markov Diffusion Kernel to derive a variant of GCN, and it can be used as a trade-off of low-pass and high-pass filter which captures the global and local contexts of each node.
    
    \item \textbf{FastGCN}~\cite{chen2018fastgcn} interprets graph convolutions as integral transforms of embedding functions under probability measures, and enhances GCN with importance sampling.
    
    \item \textbf{ClusterGCN}~\cite{chiang2019cluster} designs the batches based on efficient graph clustering algorithms, and it proposes a stochastic multi-clustering framework to improve the convergence.

    \item \textbf{GBP} ~\cite{chen2020scalable}: GBP utilizes a localized bidirectional propagation process to further improve SGC.
    
\end{itemize}

\subsection{Experiments setup}
\label{appendix:setup}
We use PyTorch~\footnote{https://github.com/pytorch} and DGL to implement the models, and we train them using Adam optimizer. To evaluate the scalability of GraphSAGE, we implement GraphSAGE via DistDGL. 
Besides, we train each model 400 epochs and terminate the training process if the validation accuracy does not improve for 20 consecutive steps. 
Note that both GraphSAGE and JKNet have three aggregators, and we choose the concatenation and mean as their aggregator, respectively, since these two aggregators perform best in most datasets. 

For GAT, the number of attention heads is fixed to 8.
For GraphSAGE, we use the results on Flickr and Reddit as reported in \cite{hamilton2017inductive} and \cite{zeng2020graphsaint}. For ClusterGCN, we use the results on Reddit as reported in \cite{chiang2019cluster} and run our own implementation on Flickr.
The hyperparameters are selected from random search. The random search was
performed over the following search space: hidden size $\in$ \{8, 16,
32, 64, 128, 256, 512\}, learning rate $\in$ \{1e-3, 5e-3, 1e-2, 5e-2, 1e-1, 2e-1\}, dropout rate $\in$ \{0.2, 0.3, 0.4,
0.5, 0.6, 0.7, 0.8, 0.85, 0.9]\},  regularization strength $\in$ \{1e-4, 5e-4,1e-3, 5e-3, 1e-2, 5e-2, 1e-1\}.
Note that both Res-GCN and JK-Net will degrade into GCN if they have only two layers, so we set their aggregation steps $\in$ [3,20] in all datasets.

\subsection{Experiment Environment and Reproduction Instructions}
\label{appendix:reproduction}
The experiments are implemented on 4 machines with 14 Intel(R) Xeon(R) CPUs (Gold 5120 @ 2.20GHz) and four NVIDIA TITAN RTX GPUs.
The code is written in Python 3.6, and the multi-objective algorithm is implemented based on OpenBox~\cite{li2021openbox}.
We use Pytorch 1.7.1 on CUDA 10.1 to train the model on GPU.

\end{document}